\documentclass{article}
\usepackage{jfrExamplee}
\usepackage{graphicx}
\usepackage{apalike}
\usepackage{setspace}

\usepackage{xcolor}
\definecolor{commentgreen}{RGB}{0, 153, 0}

\usepackage{amsthm}
\usepackage{amsmath}               
{
  \theoremstyle{plain}

}
\usepackage{amssymb}
\usepackage{mathtools}
\usepackage{graphicx}
\usepackage{tikz}
\usepackage{bm}
\usepackage{tabularx}
\usepackage[]{hyperref}
\usepackage{xcolor}
\usepackage{algorithm, algorithmicx}
\usepackage[noend]{algpseudocode}
\usepackage{pgffor}
\usepackage{etoolbox}
\usepackage{multirow}
\usepackage{ifthen}
\usepackage{subcaption}
\usepackage{xspace}

\algrenewcommand\algorithmicindent{0.9em}%
\algnewcommand{\algorithmicgoto}{\textbf{go to} line}%
\algnewcommand{\Goto}[1]{\algorithmicgoto~\ref{#1}}%

\newcommand{\BEAS}{\begin{eqnarray*}}
\newcommand{\EEAS}{\end{eqnarray*}}
\newcommand{\BEA}{\begin{eqnarray}}
\newcommand{\EEA}{\end{eqnarray}}
\newcommand{\BEQ}{\begin{equation}}
\newcommand{\EEQ}{\end{equation}}
\newcommand{\BIT}{\begin{itemize}}
\newcommand{\EIT}{\end{itemize}}
\newcommand{\BNUM}{\begin{enumerate}}
\newcommand{\ENUM}{\end{enumerate}}

\newcommand{\BA}{\begin{array}}
\newcommand{\EA}{\end{array}}

\newcommand{\reals}{\mathbb R}

{\begin{quote}}{\end{quote}}

\newcounter{oursection}

\makeatletter
\def\optimize{\@ifnextchar[{\@with}{\@without}}
\def\@with[#1]#2#3#4#5{
  \ifthenelse{\equal{#2}{}}{
    \begin{aligned}
      #3_{#4}\text{ } & #5 \\
      \mathrm{s.t.}\text{ } & #1
    \end{aligned}
  }{
    \begin{aligned}
      #2 & = & #3_{#4}\text{ } & #5 \\
         &   & \mathrm{s.t.}\text{ } & #1
    \end{aligned}
  }
}
\def\@without#1#2#3#4{
  \ifthenelse{\equal{#1}{}}{
    \begin{aligned}
      #2_{#3}\text{ } & #4
    \end{aligned}
  }{
    \begin{aligned}
      #1 & = & #2_{#3}\text{ } & #4
    \end{aligned}
  }
}
\makeatother

\definecolor{darkolivegreen}{rgb}{0.33, 0.42, 0.18}

\newcommand{\Acikmese}{A\c{c}{\i}kme\c{s}e\xspace}

\title{Long-Duration Fully Autonomous Operation of Rotorcraft UAS for
  Remote-Sensing Data Acquisition}

\author{
Danylo Malyuta \\
Autonomous Control Lab\\
University of Washington\\
Seattle, WA 98195 \\
\texttt{danylo@uw.edu} \\
\And
Christian Brommer \\
Control of Networked Systems Group \\
Alpen-Adria-Universit{\"a}t Klagenfurt \\
Klagenfurt, Austria 9020 \\
\texttt{christian.brommer@ieee.org} \\
\And
Daniel Hentzen \\
Jet Propulsion Laboratory \\
California Institute of Technology \\
Pasadena, CA 91109-8099 \\
\texttt{daniel.r.hentzen@jpl.nasa.gov} \\
\And
Thomas Stastny \\
Autonomous Systems Lab \\
ETH Z\"urich \\
Z\"urich, Switzerland 8092 \\
\texttt{thomas.stastny@mavt.ethz.ch} \\
\And
Roland Siegwart \\
Autonomous Systems Lab \\
ETH Z\"urich \\
Z\"urich, Switzerland 8092 \\
\texttt{rsiegwart@ethz.ch} \\
\And
Roland Brockers \\
Jet Propulsion Laboratory \\
California Institute of Technology \\
Pasadena, CA 91109-8099 \\
\texttt{brockers@jpl.nasa.gov}
}

\begin{document}

\maketitle

\begin{abstract}
  Recent applications of unmanned aerial systems (UAS) to precision agriculture
  have shown increased ease and efficiency in data collection at precise remote
  locations. However, further enhancement of the field requires operation over
  long periods of time, e.g. days or weeks.
  This has so far been impractical due to the limited flight times of such
  platforms and the requirement of humans in the loop for operation.
  To overcome these limitations, we propose a fully autonomous rotorcraft UAS
  that is capable of performing repeated flights for long-term observation
  missions without any human intervention.
  We address two key technologies that are critical for such a system: full
  platform autonomy to enable mission execution independently from human
  operators and the ability of vision-based precision landing on a recharging
  station for automated energy replenishment.
  High-level autonomous decision making is implemented as a hierarchy of master
  and slave state machines. Vision-based precision landing is enabled by
  estimating the landing pad's pose using a bundle of AprilTag fiducials
  configured for detection from a wide range of altitudes.
  We provide an extensive evaluation of the landing pad pose estimation accuracy
  as a function of the bundle's geometry. The functionality of the complete
  system is demonstrated through two indoor experiments with a duration of 11
  and 10.6~hours, and one outdoor experiment with a duration of 4~hours. The UAS
  executed 16, 48 and 22 flights respectively during these experiments. In the
  outdoor experiment, the ratio between flying to collect data and charging was
  1 to 10, which is similar to past work in this domain.
  All flights were fully autonomous with no human in the loop. To our best
  knowledge this is the first research publication about the long-term outdoor
  operation of a quadrotor system with no human interaction.
\end{abstract}

\section{Introduction}

\subsection{Motivation}
\label{subsec:motivation}

The Food and Agriculture Organization of the United Nations predicts that food
production will have to increase by 70 percent by the year 2050 in order to feed
a projected additional 2.2 billion people \cite{FAOFood}. Precision agriculture,
also known as smart farming, is a data-driven approach that helps to address
this challenge by using sensing technology to increase farming efficiency. Data
is collected for variables like plant and animal health, crop yields, organic
matter content, moisture, nitrogen and pH levels. To complement sensors mounted
on farming vehicles, fixed-wing and rotorcraft Unmanned Aerial Systems (UAS) can
be equipped with RGB, thermal and hyperspectral cameras and flown over fields in
order to create cost-effective and on-demand orthomosaic maps of biophysical
parameters. The market for such robots is projected to be the second-largest in
the commercial drone sector \cite{GoldmanSachs}.

Rotorcraft UAS are ideal assets for deploying sensors and instruments in a 3D
environment. Because they can hover and fly slowly, rotorcraft enable pin-point
data gathering and can be operated in dense environments such as under tree
canopies that are inaccessible to traditional fixed-wing platforms. In the
agricultural domain, an important application is the repeated acquisition of the
same target to capture changes in environmental properties such as plant water
usage over the diurnal cycle. This is of increasing importance for ecosystem
monitoring applications where accurate measurements of plant health parameters
are desired down to leaf-level resolution of individual plants.

\begin{figure}
  \center
  \includegraphics[width=0.8\columnwidth]{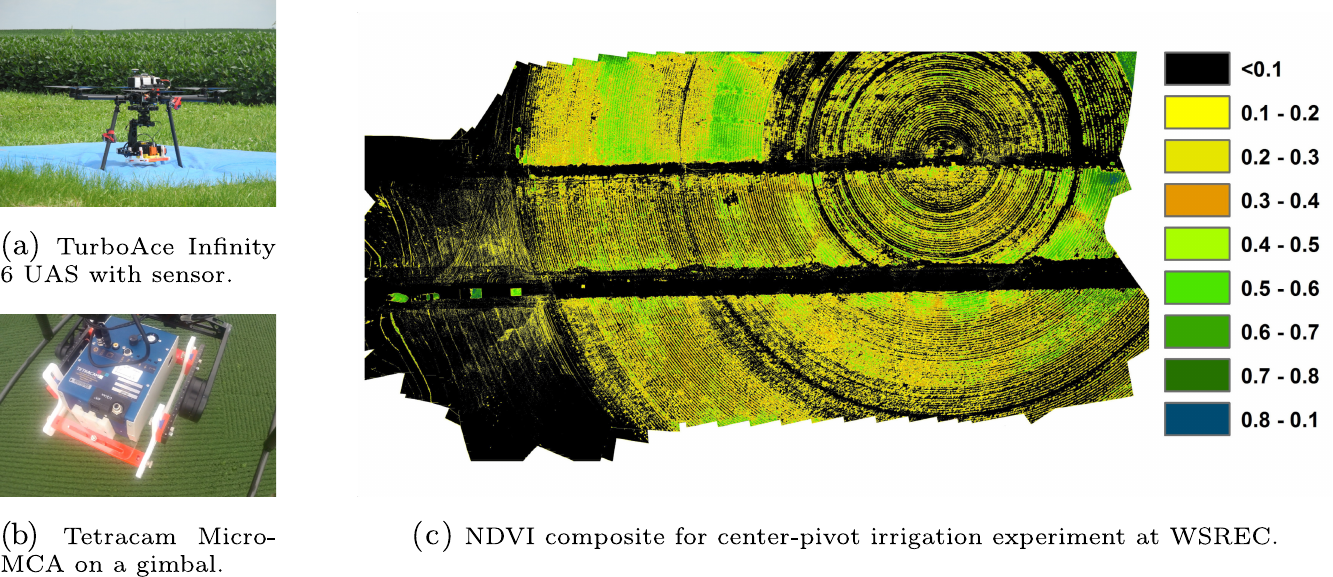}
  \caption{Example deployment of a hexacopter UAS
    (\hyperref[fig:example_application]{a}) with a hyperspectral camera
    (\hyperref[fig:example_application]{b}) to collect data for NDVI calculation with
    centimeter accuracy (\hyperref[fig:example_application]{c}). The autonomy engine
    in this paper can be deployed on such a system as a future application.}
  \label{fig:example_application}
\end{figure}

To validate the feasibility of UAS remote sensing of relevant data at high
spatial resolutions, extensive data acquisition trials were carried out in
collaboration with NASA's Jet Propulsion Laboratory environmental
scientists. Figures~\hyperref[fig:example_application]{1a} and
\hyperref[fig:example_application]{1b} show a hexarotor with a six-band visible
and near-infrared (VNIR) camera and a thermal infrared camera payload that was
used during several campaigns at the West Side Research and Extension Center
(WSREC) at Five Points, California, to capture the diurnal cycle of plant water
usage. Figure~\hyperref[fig:example_application]{1c} illustrates a stitched map
of the normalized difference vegetation index (NDVI) as an example data product
of one flight. The map's pixel resolution of approximately 5~cm is beyond what
can be currently achieved with orbital or traditional airborne instruments. The
autonomy engine developed in this paper can be deployed on such a quadrotor as a
potential future application.

\subsection{Contributions}

Because the typical flight endurance of a quadrotor system with payload is
relatively short (typically 10-20~min), long-term operation currently requires
humans in the loop for energy replenishment (e.g. battery charging). This makes
continuous observations over long periods of time impractical. To solve this
issue, we propose a system that is capable of fully autonomous repeated mission
execution, including automated recharging to replenish its batteries after a
data collection flight is completed.

Our main contribution is the development and outdoor deployment of a high-level
autonomy engine that is executed on-board the UAS and which implements the
capability of robust fully autonomous repeated data acquisition. This includes
flight control for take-off and landing, GPS-inertial state estimation, system
health monitoring including the execution of emergency behaviors, recharging and
mission data handling. Once the autonomy engine is started, the system has the
logic and software-hardware interfaces necessary to operate indefinitely without
humans in the loop.

The system is novel by being \textit{human-independent} and \textit{operating
  outdoors}. To the best of our knowledge, there have been no publications to
date on quadrotor-type UAS platforms capable of autonomous recharging which have
been flown outdoors for extended periods of time. Hence, prior work that
demonstrates UAS recharging has relied on the idealistic state estimation
accuracy offered by indoor motion capture systems. Since GPS-only localization
is not sufficiently accurate for precision landing on a charging pad, our work
augments the sensor suite with a monocular camera. In this context, we present
an extensive analysis seeking to find a landing pad fiducial marker layout that
maximizes landing accuracy.

\subsection{Notation}

The following notation is used throughout the report. Scalars are lowercase
(e.g. $a$), vectors are lowercase bold (e.g. $\bm{a}$) and matrices are
uppercase (e.g. $A$). The position of frame $\{\text{b}\}$ in frame
$\{\text{a}\}$ is written $\bm{p}_{\text{a}}^{\text{b}}$. The quaternion active
rotation from $\{\text{a}\}$ to $\{\text{b}\}$ is written
$\bm{q}_{\text{a}}^{\text{b}}$ and the corresponding rotation matrix is
$C_{\bm{q}_{\text{a}}^{\text{b}}}$. The homogeneous transformation matrix
$T_{\text{ab}}$ takes homogeneous vectors in $\{\text{b}\}$ to
$\{\text{a}\}$. Raw measurement quantities carry a hat,
e.g. $\hat{\bm{p}}_{\text{a}}^{\text{b}}$, while filtered quantities carry a
tilde, e.g. $\tilde{\bm{p}}_{\text{a}}^{\text{b}}$.

\subsection{Outline}

This paper is organized as follows. Section~\ref{sec:related_work} surveys
existing work on long-duration autonomy and vision-based landing, the two key
enabling technologies for our system. The hardware and software architectures
are then described in
Section~\ref{sec:system_overview}. Sections~\ref{sec:navigation},
\ref{sec:guidance} and \ref{sec:control} detail the navigation, guidance and
control subsystems respectively which enable the UAS to execute flight
tasks. The autonomy engine itself is presented in
Section~\ref{sec:autonomy_engine}. Field tests with the system in environments
of increasing complexity, up to real world conditions, are then given in
Section~\ref{sec:results}, followed by a discussion of future work and
conclusion in Sections~\ref{sec:discussion} and \ref{sec:conclusion}.

\section{Related Work}
\label{sec:related_work}

In this section we review related work in our two main areas of contribution,
the ability to replenish energy for long-duration autonomy and vision-based
precision landing.

\subsection{Long-duration Autonomy}

Several rotorcraft UAS that are capable of autonomous, long-duration mission
execution
in benign indoor (VICON) environments have previously appeared in literature.
Focusing on the recharging solution to extend individual platform flight time
and a multi-agent scheme for constant operation, impressive operation times have
been demonstrated
(\cite{Valenti2007MissionHM}: 24 h single vehicle experiment;
\cite{Mulgaonkar_2014}: 9.5 h with multiple vehicles).
Recharging methods vary from wireless charging \cite{Aldhaher_2017} to contact-based charging pads \cite{Mulgaonkar_2014} to battery swap systems \cite{Toksoz_2011,Suzuki_2012}.
While wireless charging offers the most flexibility since no physical contact
has to be made, charging currents are low resulting in excessively long charging
times and are hence not an option for quick redeployment. However, interesting
results have been shown in \cite{Aldhaher_2017} demonstrating wireless powering
of a 35 g/10 W micro drone in hover flight.  On the other end of the spectrum,
battery swap systems offer immediate redeployment of a UAS, but require
sophisticated mechanisms to be able to hot swap a battery, and a pool of charged
batteries that are readily available.  This makes such systems less attractive
for maintenance free and cost effective long-term operation.

\subsection{Vision-based Landing}
\label{subsec:related_work_vision_based_landing}

Our work uses a downward-facing monocular camera to estimate the pose of the
landing pad in the world frame using AprilTag visual fiducial markers
\cite{wang2016iros}. We believe that a monocular camera is the cheapest, most
lightweight and power efficient sensor choice. Alternatively, GPS and RTK-GPS
systems suffer from precision degradation and signal loss in occluded urban or
canyon environments \cite{Shaogang2016}. Laser range finder systems are heavy
and consume considerable amounts of energy. Last but not least, stereo camera
setups have limited range as a function of their baseline versus vehicle-to-pad
distance.

Several landing approaches exist in literature for labeled and unlabeled landing
sites. \cite{Yang2013} present a monocular visual landing method based on
estimating the 6 DOF pose of a circled H marker. The same authors extend this
work to enable autonomous landing site search by using a scale-corrected PTAM
algorithm \cite{Klein2007} and relax the landing site structure to an arbitrary
but feature-rich image that is matched using the ORB algorithm
\cite{Rublee}. \cite{Forster,Forstera} use SVO to estimate motion from a
downfacing monocular camera to build a probabilistic 2 dimensional elevation map
of unstructured terrain and to detect safe landing spots based on a score
function. \cite{Brockers2011,Brockers2012} develop a fully self-contained visual
landing navigation pipeline using a single camera. With application to landing
in urban environments, the planar character of a rooftop is leveraged to perform
landing site detection via a planar homography decomposition using RANSAC to
distinguish the ground and elevated landing surface planes.

However, landing algorithms which build elevation maps and perform generic
landing spot detection trade their robustness for the ability to land in
unlabeled environments. By placing visual fiducial markers at or near the
landing zone, one can use visual fiducial detector algorithms to estimate the
landing pad pose more reliably, i.e. with less false positives and negatives.

Currently one of the most popular visual fiducial detectors and patterns is the
AprilTag algorithm \cite{olson2011tags} which is renowned for its speed,
robustness and extremely low false positive detection rates. The algorithm was
updated by \cite{wang2016iros} to further improve computational efficiency and
to enable the detection of smaller tags. AprilTag has been applied multiple
times for MAV landing
\cite{Brommer2018,Ling2014,Kyristsis2016,Borowczyk,Chaves2015}. Similar visual
fiducials are seen around the landing zones of both Amazon and Google delivery
drones \cite{Amazon2016,Vincent2017} as well as of some surveying drones
\cite{Wingtra}. Our approach is the same as that of \cite{Brommer2018}. We
improve over \cite{Ling2014,Kyristsis2016} by using a bundle of several tags for
improved landing pad pose measurement accuracy. \cite{Borowczyk,Chaves2015}
appear to also use a tag bundle, however they deal with a moving landing
platform and feed individual tag pose detections into a Kalman filter. Our
approach is to use a perspective-$n$-point solution to obtain a single pose
measurement using all tags at once.

After obtaining raw tag pose estimates from the AprilTag detector, our approach
uses a recursive least squares (RLS) filter to obtain a common tag bundle pose
estimate. A relevant previous work on this topic is \cite{Nissler2016} in which
a particle filter is applied to raw tag detections and RLS is compared to RANSAC
for bundle pose estimation. Because the demonstrated accuracy improvement is
marginal (about 1~cm in mean position and negligible in mean attitude) and
RANSAC has disadvantages like ad hoc threshold settings and a non-deterministic
runtime, we prefer RLS for its simplicity. Nevertheless, RANSAC and its more
advanced variations \cite{Hast2013} can be substituted into our
implementation. Other work investigated fusing tag measurements in RGB space
with a depth component \cite{Jin2017} with impressive gains in accuracy. One can
imagine this approach benefiting landing accuracy at low altitudes, however
downward-facing stereo camera mounting on drones raises several concerns like
weight, vibration effects and space availability.

Our approach, however, is not limited to AprilTags, which can be substituted or
combined with other markers for specific applications. A vast number of markers
is available targeting different use-cases
\cite{Shaogang2016,Yang2013,Fiala,olson2011tags}. Patterns include letters,
circles, concentric circles and/or polygons, letters, 2 dimensional barcodes and
special patterns based on topological \cite{Bencina}, detection range maximizing
\cite{Xu2011} and blurring/occlusion robustness considerations
\cite{Bergamasco2016}.

\section{System Overview}
\label{sec:system_overview}

\begin{figure}
  \centering
  \includegraphics[width=0.6\columnwidth]{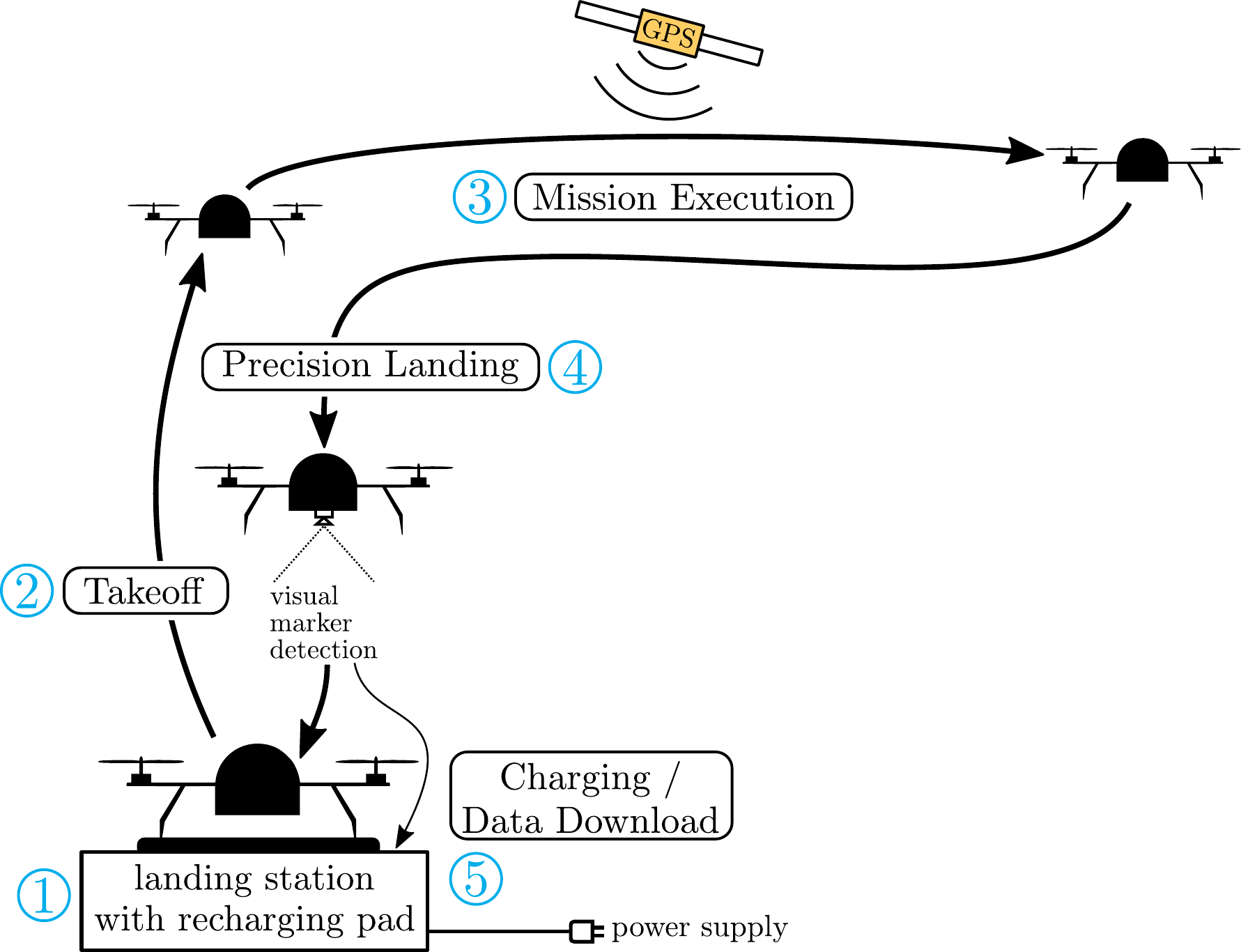}
  \caption{Autonomous UAS data acquisition cycle. 1: the vehicle is placed on
    the landing station to start the operation. 2: takeoff to safe altitude. 3:
    mission execution. 4: vision-based precision landing. 5: recharging and data
    downloading.}
  \label{fig:mission_scheme}
\end{figure}

\begin{figure}
  \centering
  \includegraphics[width=0.7\columnwidth]{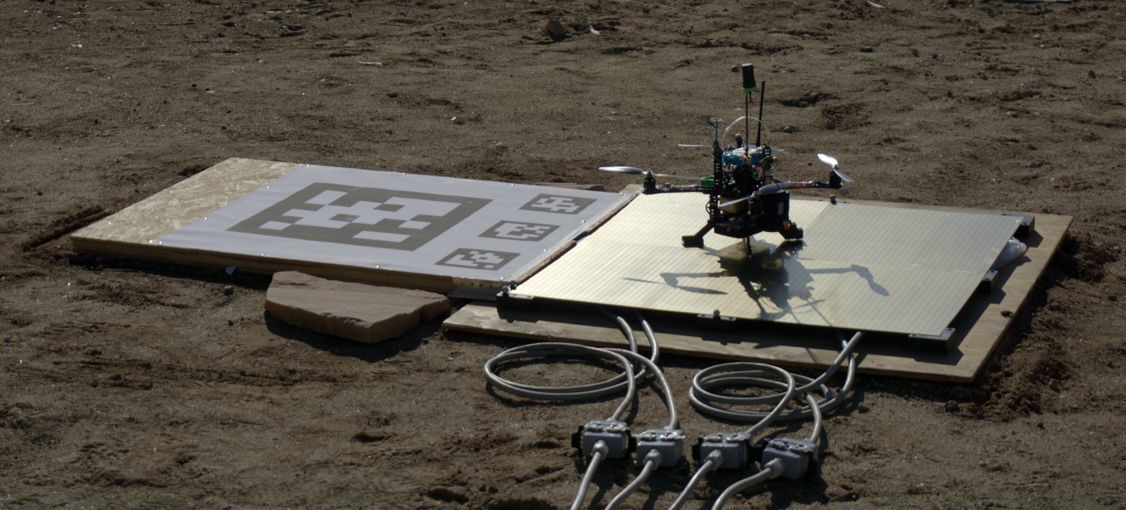}
  \caption{Landing station with visual bundle (left), charging pad (right), and
    UAS. The recharging process starts automatically when the quadrotor makes
    contact with the 90$\times$90~cm charging surface.}
  \label{fig:landing_pad_with_QR}
\end{figure}

Our system consists of two major components: the aerial vehicle with its
on-board autonomy software (autonomy engine) and a landing station which
integrates a 
charging pad, a base station computer for downloading data,
and a wireless router for communication with the UAS while landed. To deploy the
system, a user connects the landing station to a power outlet and places the UAS
on top of the landing surface as shown in Figures~\ref{fig:mission_scheme} and
\ref{fig:landing_pad_with_QR}. After providing a waypoint based mission profile,
a start command commences the operation of the system. All actions of the UAS
from then on are fully autonomous with no human involvement.
The autonomy engine implements various mission specific behaviors to guide the
vehicle, including health monitoring and emergency response that are explained
in Section~\ref{sec:autonomy_engine}.

\subsection{Hardware Architecture}
\label{subsec:hardware_architecture}

\begin{figure}
  \centering
  \includegraphics[width=0.7\columnwidth]{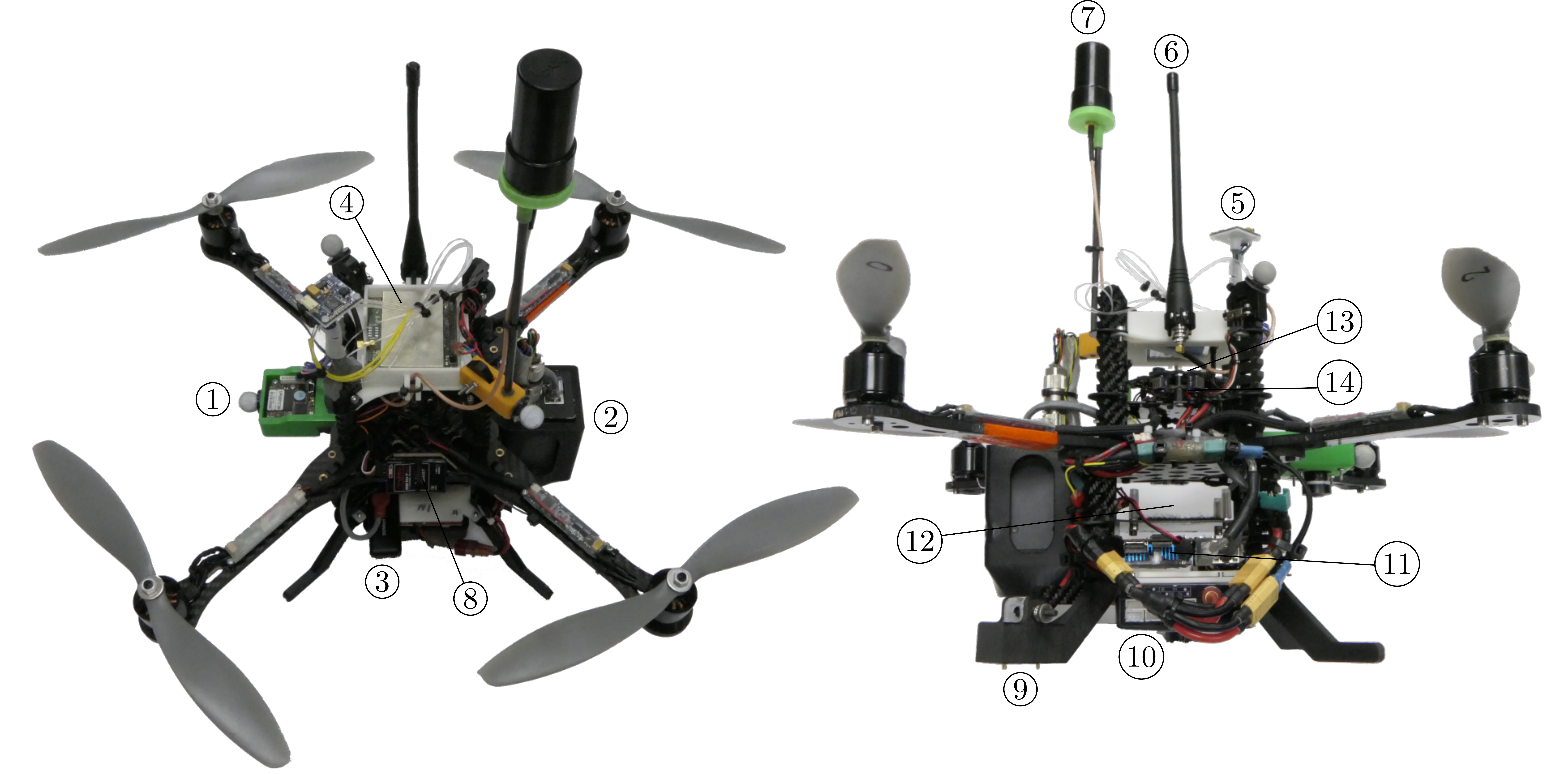}
  \vspace{2mm}
  
  \begin{tabular}{r|l}
    1 & (Navigation camera) USB 2.0 mvBlueFOX–MLC200wG downfacing camera with a
        100$^{\circ}$ FOV lens \\
    2 & (Payload) FLIR Ax5-series thermal camera \\
    3 & 5 GHz dual band WiFi module \\
    4 & Trimble BD930-UHF GPS receiver \\
    5 & 3-axis magnetometer \\
    6 & UHF antenna \\
    7 & GPS antenna \\
    8 & RC receiver \\
    9 & Leg charging contacts \\
    10 & CRC C6S LiPo balance charger \\
    11 & Odroid XU4 navigation computer \\
    12 & 3S1P 6200 mAh LiPo battery holder \\
    13 & AscTec AutoPilot Board \\
    14 & AscTec Power Board
  \end{tabular}
  \caption{Hardware elements on our test UAS (modified AscTec Pelican).}
  \label{fig:quad_labeled}
\end{figure}

\begin{figure}
  \centering
  \includegraphics[width=0.8\columnwidth]{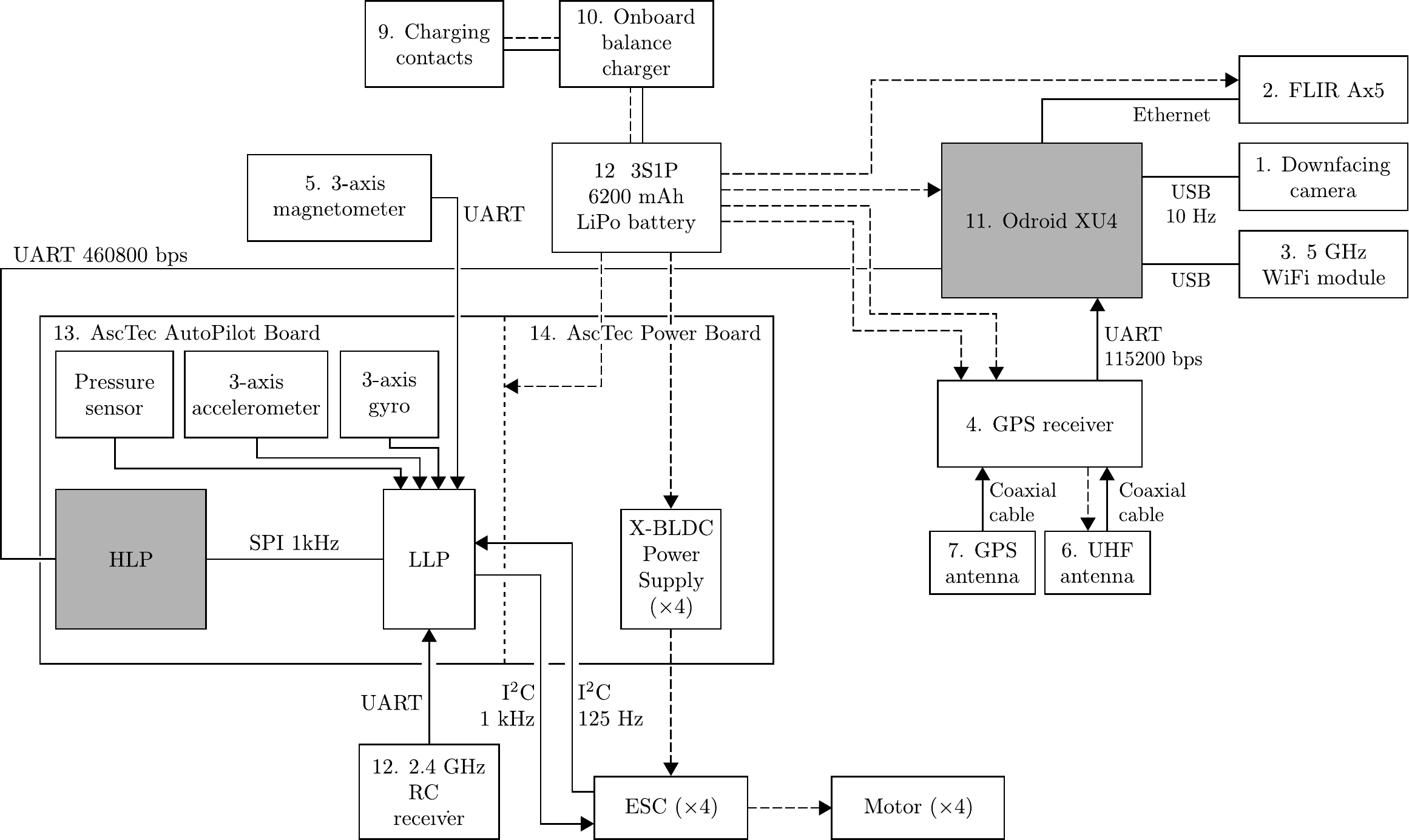}
  \caption{On-board hardware architecture and information flow. Solid and dashed
    lines represent information and energy flows respectively. Communication
    protocols are labelled where known.}
  \label{fig:hardware_architecture_quadrotor}
\end{figure}

Figure~\ref{fig:quad_labeled} illustrates the main hardware elements of our test
UAS. Figure~\ref{fig:hardware_architecture_quadrotor} displays the quadrotor
on-board hardware architecture. Solid lines represent information flow and are
labelled whenever the communication protocol is known, and dashed lines
represent energy flow. The gray blocks in
Figure~\ref{fig:hardware_architecture_quadrotor} label the components that host
the flight software developed in this work.

To minimize recharging time, we use a contact-based, commercially available
charging solution \cite{skysense_2017} to provide ample charging current to the
on-board charging electronics. The charging pad consists of an array of
individual contact patches that cover a $90\times 90$ cm flat area. Charging is
triggered by the charging pad electronics once contact to the charger on-board
the UAS is made via leg contacts. 
Charging status is monitored online by the autonomy engine, which prepares for
the next flight once the battery is sufficiently charged.
While landed, the vehicle connects to the base station computer via WiFi to
downlink mission data and to receive eventually updated mission plans
(e.g. mission-end command from a user).

\subsection{Software Architecture}

\begin{figure}
  \centering
  \includegraphics[width=0.9\textwidth]{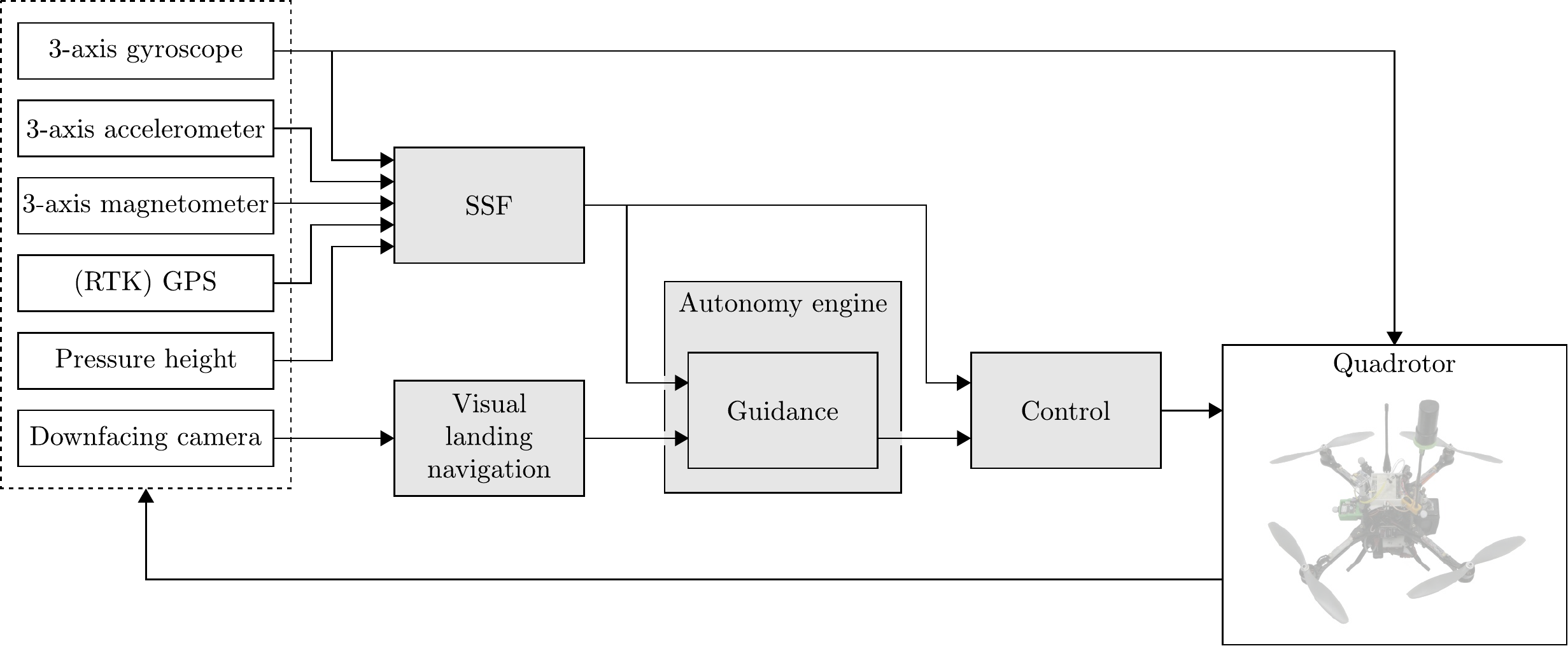}
  \caption{Sensor and software system block diagram. Gray blocks are part of the
    software subsystem.}
  \label{fig:system_information_flow}
\end{figure}

Figure~\ref{fig:system_information_flow} illustrates the full autonomy system as
an information flow diagram. Sensors feed information to the Single Sensor
Fusion (SSF) state estimator \cite{Weiss2012}, modified to handle multiple
sensors, and to the visual landing navigation subsystems. These supply the
control, guidance and autonomy algorithms with navigational information
necessary for stabilization, trajectory generation and state machine
transitions. The control subsystem is then responsible for trajectory tracking
and desired motor speed computation.

\subsection{Mission Architecture}

The UAS executes a mission profile depicted in
Figure~\ref{fig:mission_scheme}. Before each take-off, the system initializes
the on-board state estimator and passes a series of pre-flight checks that
include testing for an adequate battery voltage level and motor nominal
performance.
The vehicle then performs a take-off maneuver by issuing a vertical velocity
command to its low-level velocity controller to climb to an initial altitude.
Once a safe altitude is reached, the mission execution module within the
autonomy engine takes over.  A mission is defined as a set of waypoints
connected by individual trajectory segments that are calculated using polynomial
trajectories following the approach of \cite{Mellinger2011,Richter2013}.
A trajectory tracking module plays the mission trajectory forward in time by
issuing position references to the low-level position controller. During mission
execution, a system health observer monitors all critical components of the UAS
and issues emergency messages to the autonomy engine to implement fail-safe
behaviors (e.g. return-to-home, emergency-landing on low battery or state
estimation failure).

After the mission is completed, the UAS returns to the vicinity of the landing
station using the recorded GPS position of the take-off location. Our
vision-based landing navigation algorithm then detects the landing station
AprilTag fiducials with its downfacing camera. Once the landing location is
accurately detected, a landing maneuver is executed by first hovering over the
landing location and then issuing a trajectory with a fixed descent velocity
that passes through the center of the charging pad surface. Touchdown is
triggered by detecting zero velocity below a minimum altitude threshold.

\section{Navigation}
\label{sec:navigation}

The navigation subsystem is responsible for estimating the UAS state and the
landing pad pose in the world frame. The landing pad pose estimation is designed
to provide a sufficiently accurate pose estimate to be able to land on a
$90\times 90$~cm surface. The vehicle pose estimator is designed to work in an
outdoor environment such as a crop field.

\subsection{Visual Landing Navigation}
\label{subsec:visual_landing_navigation}

\begin{figure}
  \centering
  \includegraphics[width=1\textwidth]{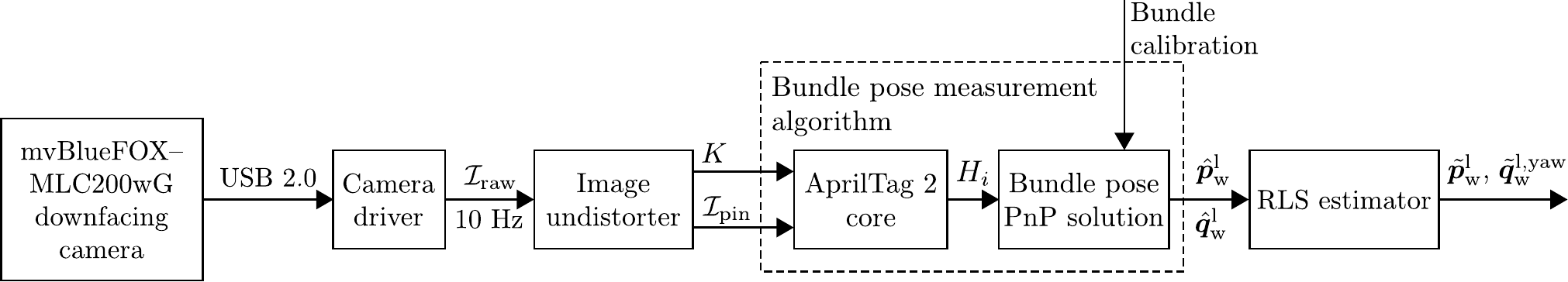}
  \caption{Vision-based landing navigation block diagram. A downfacing camera
    feeds images to a radial undistorter. AprilTag markers are detected in the
    resulting images and their pose is jointly estimated via a
    perspective-$n$-point solution. An RLS estimator smoothens the resulting
    signal based on a current estimate of the measurement variance.}
  \label{fig:visual_navigation_block_diagram}
\end{figure}

\begin{figure}
  \centering
  \includegraphics[width=0.6\columnwidth]{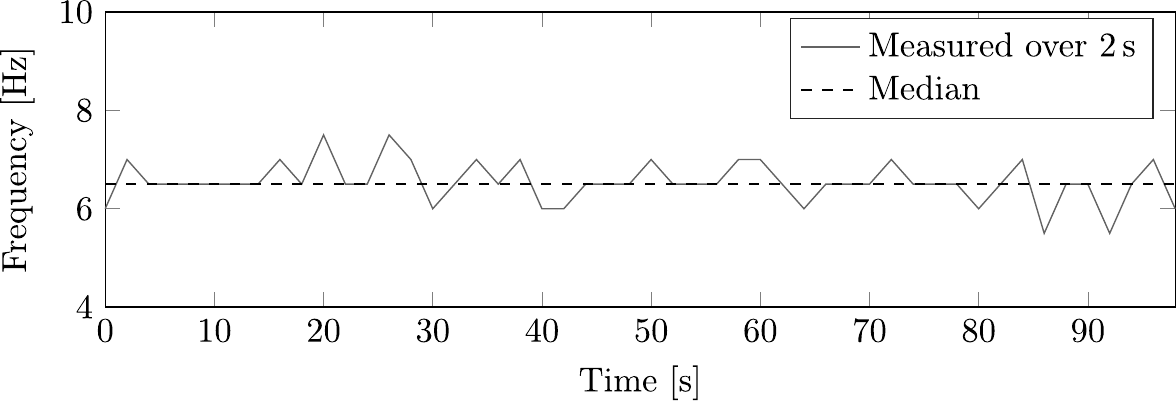}
  \caption{Landing bundle pose measurement frequency. The median is 6.5 Hz.}
  \label{fig:at_frequency}
\end{figure}

Figure~\ref{fig:visual_navigation_block_diagram} details our landing site
detection algorithm. The downfacing camera feeds a distorted 752$\times$480~px
grayscale image, $\mathcal I_{\text{raw}}$, into the image undistorter. The
latter removes radial distortion in the image via the ideal fish-eye lens model
\cite{Devernay}, producing the un-distorted image $\mathcal I_{\text{pin}}$
assumed to be produced by a pinhole camera with calibration matrix
$K\in\reals^{3\times 3}$. The unmodified AprilTag~2 implementation
\cite{APRILLaboratory2015} identifies AprilTag markers in
$\mathcal I_{\text{pin}}$ and for each tag with ID $i$ produces a homography
matrix $H_i\in\reals^{3\times 3}$. Assuming that at least one tag is detected, a
perspective-$n$-point solver \cite{OpenCV} produces an AprilTag bundle pose
measurement $\hat{\bm{p}}_{\text{w}}^{\text{l}}$,
$\hat{\bm{q}}_{\text{w}}^{\text{l}}$ of the landing pad in the world frame. This
measurement is passed to a recursive least squares (RLS) estimator to output
$\tilde{\bm{p}}_{\text{w}}^{\text{l}}$ and
$\tilde{\bm{q}}_{\text{w}}^{\text{l,yaw}}$ where the latter is a pure yaw
quaternion because we assume the landing pad to be level. The RLS outputs are
then used for landing navigation. The algorithm is executed on a single core of
an Odroid XU4, except for the AprilTag 2 detector which runs on two cores, and
achieves a measurement frequency of approximately 7 Hz as shown in
Figure~\ref{fig:at_frequency}.

Estimation of the landing pad pose using a tag bundle has several
advantages. First, by providing more points to compute the homography matrix,
the landing pad pose measurement becomes more accurate and robust to effects
like tag corner detection error. Furthermore, a bundle can be composed of larger
and smaller tags such that it is visible from a wide range of altitudes (see
Section~\ref{subsubsec:optimal_bundle_layout}).

\subsubsection{Bundle Calibration}
\label{subsubsec:calibration}

\begin{figure}
  \centering
  \includegraphics[width=0.4\textwidth]{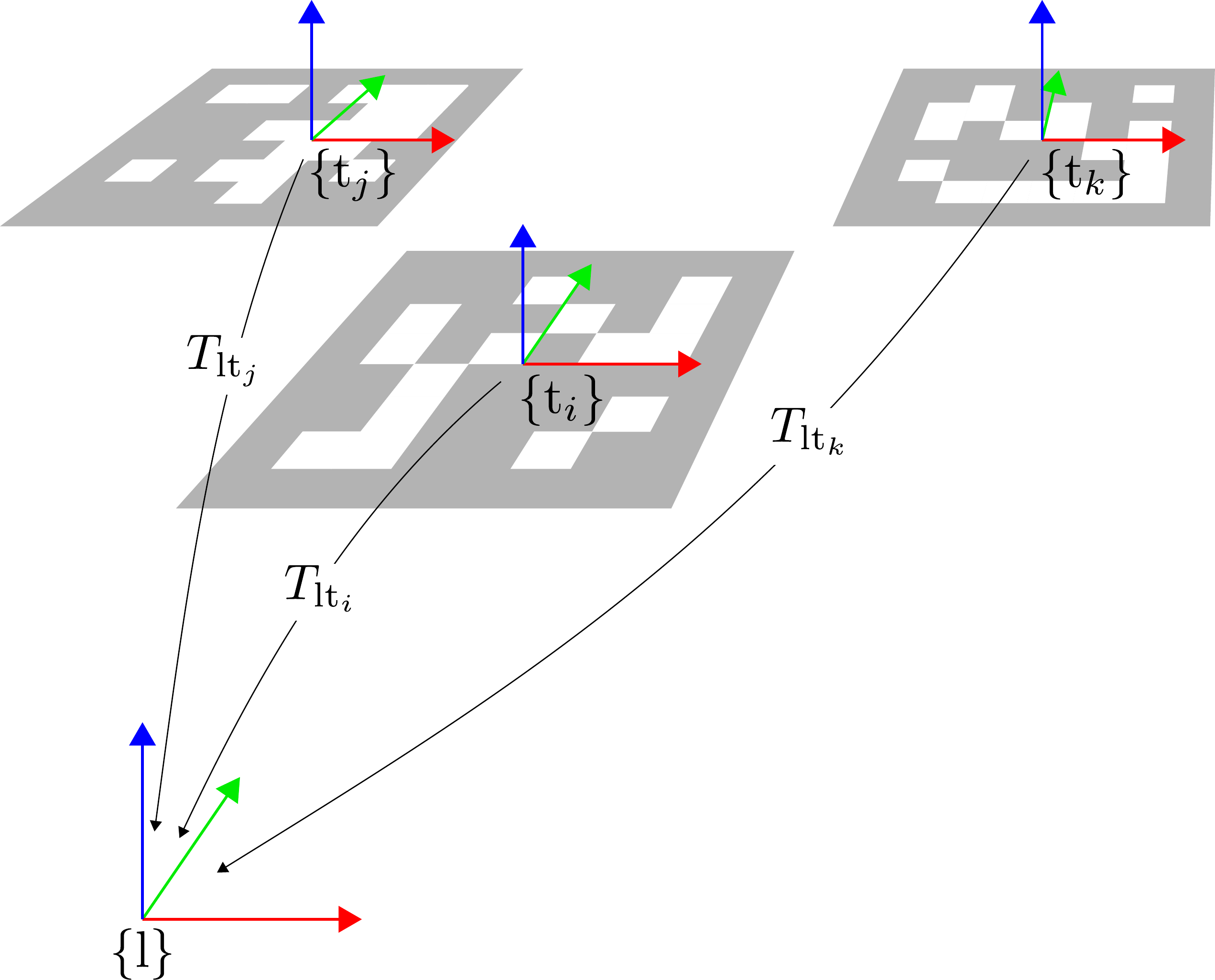}
  \caption{Rigid body transforms from each tag's local frame to the landing pad
    frame. Generic tags $i$, $j$ and $k$ are shown.}
  \label{fig:tag_pad_rigid_transform}
\end{figure}

To compute $\hat{\bm{p}}_{\text{w}}^{\text{l}}$ and
$\hat{\bm{q}}_{\text{w}}^{\text{l}}$, the transformation matrix from each tag to
the landing pad frame must be known. This is shown in
Figure~\ref{fig:tag_pad_rigid_transform} and the computation of these transforms
is called landing bundle calibration. Because the landing pad pose directly
affects the desired UAS landing pose (see
Section~\ref{subsubsec:landing_alignment}), calibration is crucial for
successful system operation.

\begin{figure}
  \centering
  \includegraphics[width=0.3\textwidth]{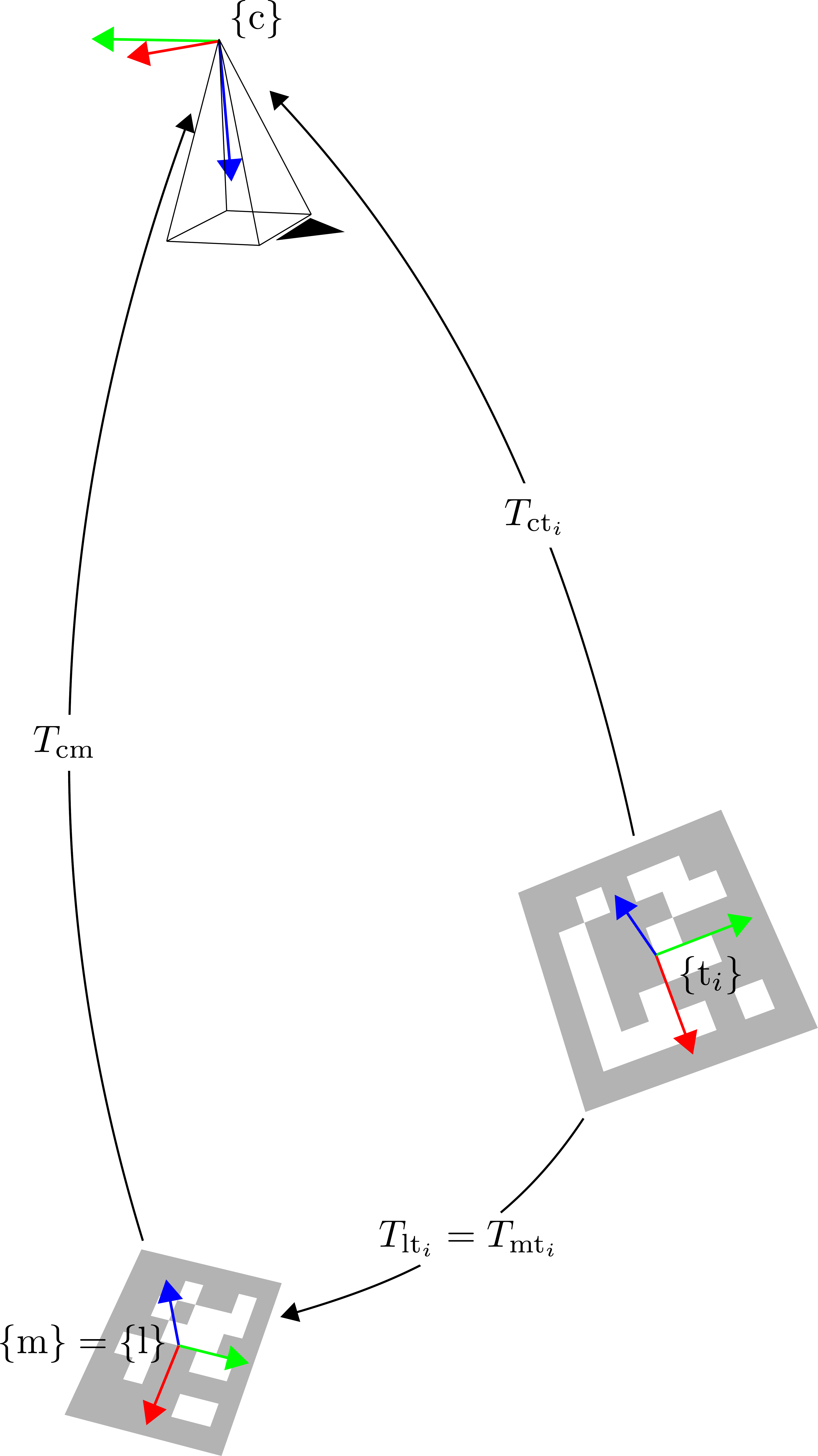}
  \caption{Rigid transform triad formed during the calibration process between
    the camera frame, the master tag/landing pad frame and the $i$-th tag
    frame.}
  \label{fig:calibration_transforms}
\end{figure}

Our calibration approach is to place the UAS downfacing camera with the landing
bundle visible, forming a rigid body transform triad as shown in
Figure~\ref{fig:calibration_transforms}. For the calibration process, a
``master'' AprilTag marker is placed at the same position and yaw as desired for
the UAS and the downfacing camera respectively during landing (see
Figure~\ref{fig:pad_alignment_position}). The unknown calibration transform
$T_{\text{lt}_i}$ is given by:
\begin{equation}
  \label{eq:T_lti_triangle_solution}
  T_{\mathrm{lt}_i} = T_{\mathrm{cm}}^{-1}T_{\mathrm{ct}_i}.
\end{equation}

However, every standalone tag pose measurement contains error due to effects
from image blur, pixelation, etc. The result of single-measurement computation
(\ref{eq:T_lti_triangle_solution}) is therefore not reliable. For this reason, a
statistical approach is taken and a sequence of several hundred
$\hat T_{\text{ct}_i}$ and $\hat T_{\text{cm}}$ is collected. These measurements
are then combined into a geometric median position and a mean quaternion
attitude of each tag relative to the landing pad frame. Computational details
are provided in \cite{Malyuta2018}.

\subsubsection{Landing Alignment}
\label{subsubsec:landing_alignment}

\begin{figure}
  \centering
  \includegraphics[width=0.7\textwidth]{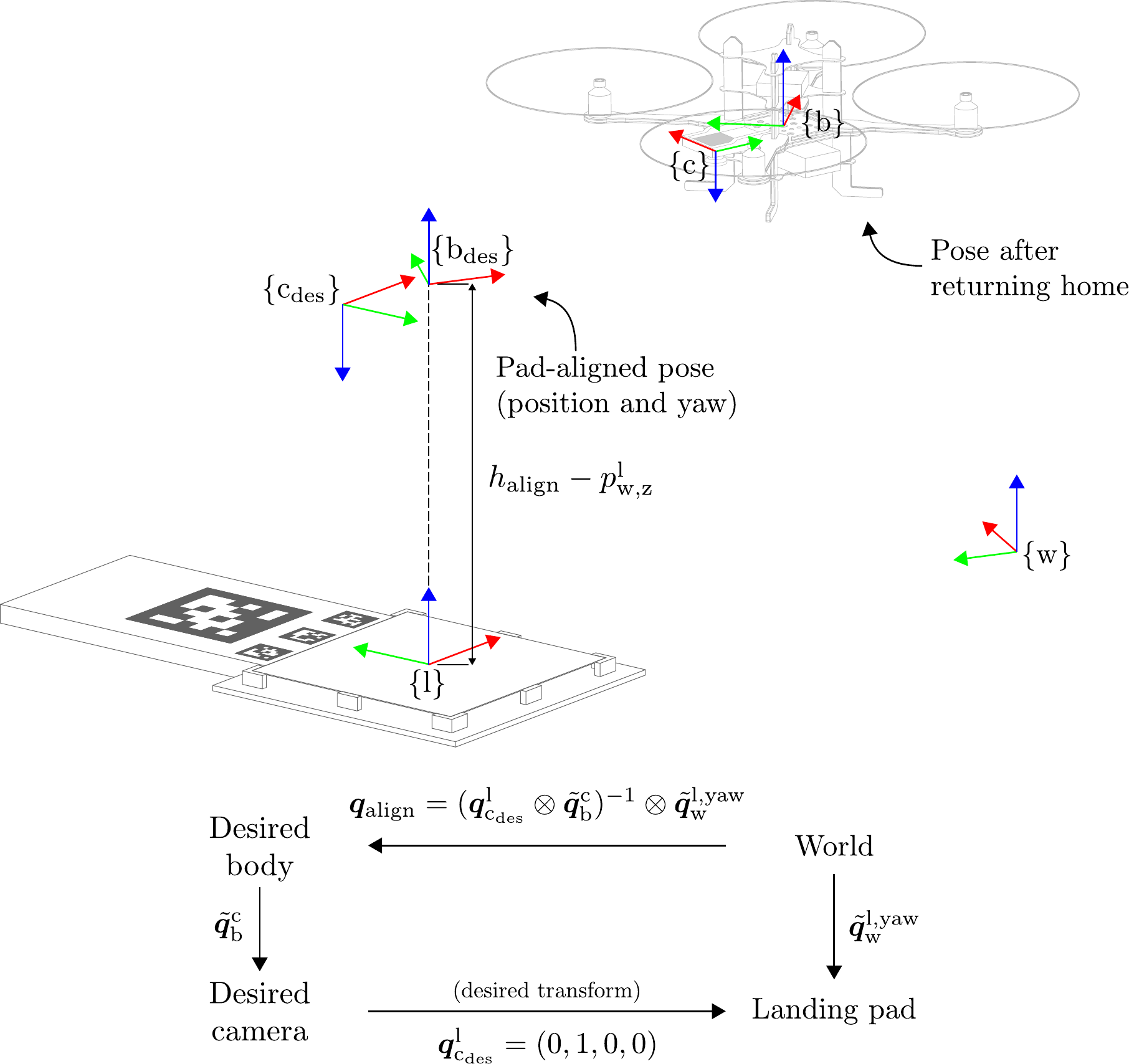}
  \caption{Frame setup for pad-aligned body frame pose computation during
    the initial part of the landing phase.}
  \label{fig:pad_alignment_position}
\end{figure}

The UAS can align itself with the landing pad by using landing pad pose estimate
$\tilde{\bm{p}}_{\text{w}}^{\text{l}}$ and
$\tilde{\bm{q}}_{\text{w}}^{\text{l,yaw}}$. For this purpose, consider the
alignment scenario depicted in Figure~\ref{fig:pad_alignment_position}. The
desired attitude of the camera is to look straight down at the landing pad
frame, hence $\bm{q}_{\text{c}_{\text{des}}}^{\text{l}}$ corresponds to a
180$^\circ$ rotation about the $x$-axis. As shown in
Figure~\ref{fig:pad_alignment_position}, we have all the elements necessary to
compute the desired body pose such that the UAS is center-aligned with the
landing frame and the camera is facing down with a yaw implicitly defined by the
master tag during the calibration step of Section~\ref{subsubsec:calibration}:
\begin{align}
  \label{eq:calibration_position}
  \bm{q}_{\text{align}}
  &= (\bm{q}_{\text{c}_{\text{des}}}^{\text{l}}\otimes \tilde{\bm{q}}_{\text{b}}^{\text{c}})^{-1}\otimes\tilde{\bm{q}}_{\text{w}}^{\text{l,yaw}}, \\
  \label{eq:calibration_attitude}
  \bm{p}_{\text{align}}
  &= \tilde{\bm{p}}_{\text{w}}^{\text{l}}+(h_{\text{align}}-
    \tilde{p}_{\text{w,z}}^{\text{l}})\bm{e}_{\text{z}},
\end{align}
where $h_{\text{align}}$ is the desired altitude above the landing pad prior to
final descent and $\bm{e}_{\text{z}}=(0,0,1)$. Once aligned, the landing
autopilot descends the UAS until touchdown as described in
Section~\ref{subsec:landing_autopilot}.

\subsection{Vehicle Pose Estimation}
\label{subsec:vehicle_pose_estimation}

The state estimation subsystem computes the vehicle pose to be used for guidance
and control. For this purpose the SSF algorithm \cite{Weiss2012} is extended to
fuse multiple sensors. In particular, we use an IMU, a GPS for position and
velocity, a pressure sensor for improved height estimation, and a 3-axis
magnetometer for yaw observability during hover and constant velocity flight.
We denote by $\{\text{b}\}$, $\{\text{g}\}$, $\{\text{m}\}$
and $\{\text{p}\}$ the body (i.e. IMU), GPS, magnetometer and pressure sensor
frames respectively and use the following state vector:
\begin{equation}
  \label{eq:ssf_state}
  \bm{x}_{\text{state}} = (\underbrace{\bm{p}_{\text{w}}^{\text{b}},
    \bm{v}_{\text{w}}^{\text{b}},\bm{q}_{\text{w}}^{\text{b}},\bm{b}_\omega,\bm{b}_{\text{a}}}_{\text{Core
      state}},
  \bm{p}_{\text{b}}^{\text{g}},\bm{q}_{\text{b}}^{\text{m}},\bm{m}_{\text{w}},b_{\text{p}},\bm{p}_{\text{b}}^{\text{p}}
  ),
\end{equation}
where $\bm{b}_\omega\in\reals^3$, $\bm{b}_{\text{a}}\in\reals^3$ and
$b_{\text{p}}\in\reals$ are the gyroscope, accelerometer and pressure
sensor biases and $\bm{m}_{\text{w}}\in\reals^3$ is a normalized magnetic field
vector. The core state portion of (\ref{eq:ssf_state}) together with the raw
gyroscope body rate measurement is used for control and guidance. During the
landing phase, the landing pad pose estimate
$\tilde{\bm{p}}_{\text{w}}^{\text{l}}$ and
$\tilde{\bm{q}}_{\text{w}}^{\text{l,yaw}}$ is used to re-localize the landing
pad in the world frame. This is necessary due to drift in the GPS sensor during
long flights.

\section{Guidance}
\label{sec:guidance}

The guidance subsystem is responsible for generating control reference signals
with the aim of following a specific trajectory or arriving at a specific
destination \cite{Grewal2007}. This section describes the guidance philosophy
and main aspects, while further computational details are provided in
\cite{Malyuta2018}.

The guidance subsystem is designed with the objectives of being computationally
lightweight and capable of generating predictable trajectories that are
dynamically feasible and that enable the UAS to explore an obstacle-free 3D
volume via point-to-point constant velocity segments. Among existing methods
that minimize snap \cite{Mellinger2011,Richter2013,Burri2015,DeAlmeida2017},
jerk \cite{Mueller2013,Mueller2015} and energy
\cite{Acikmese2007,Blackmore2010}, our approach is to compute the trajectories
as fully constrained 9-th order polynomials that are smooth up to snap at their
endpoints. This avoids online optimization and, in our experience, makes the
planned trajectory much more readily predictable than the other approaches.

\subsection{Trajectory Segments}

\begin{figure}
  \centering
  \includegraphics[width=0.5\columnwidth]{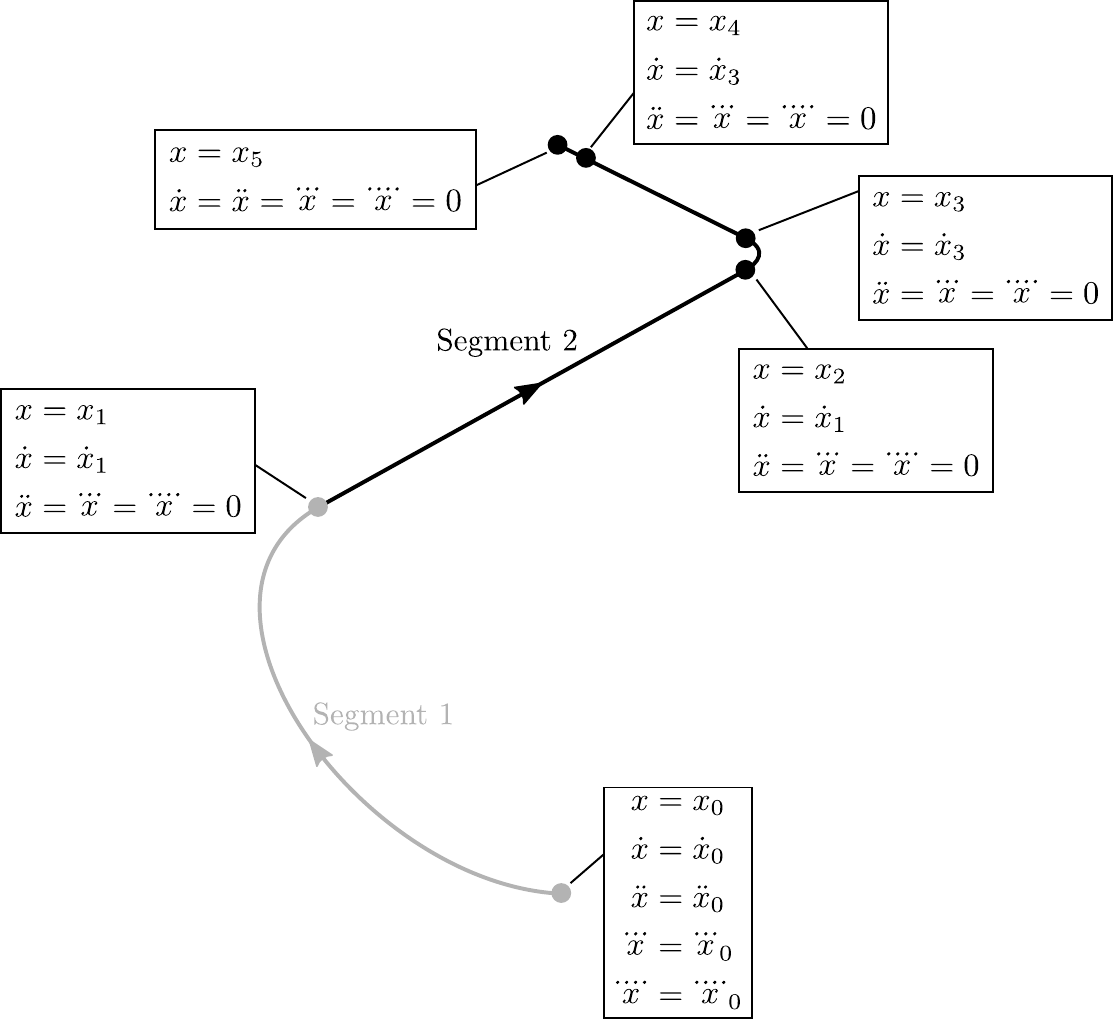}
  \caption{Illustration of a trajectory and polynomial interpolation points. All
    quadrotor trajectories are composed of sequences of either segment 1 or 2.}
  \label{fig:trajectory_illustration}
\end{figure}

\begin{figure}
  \centering
  \includegraphics[width=0.6\textwidth]{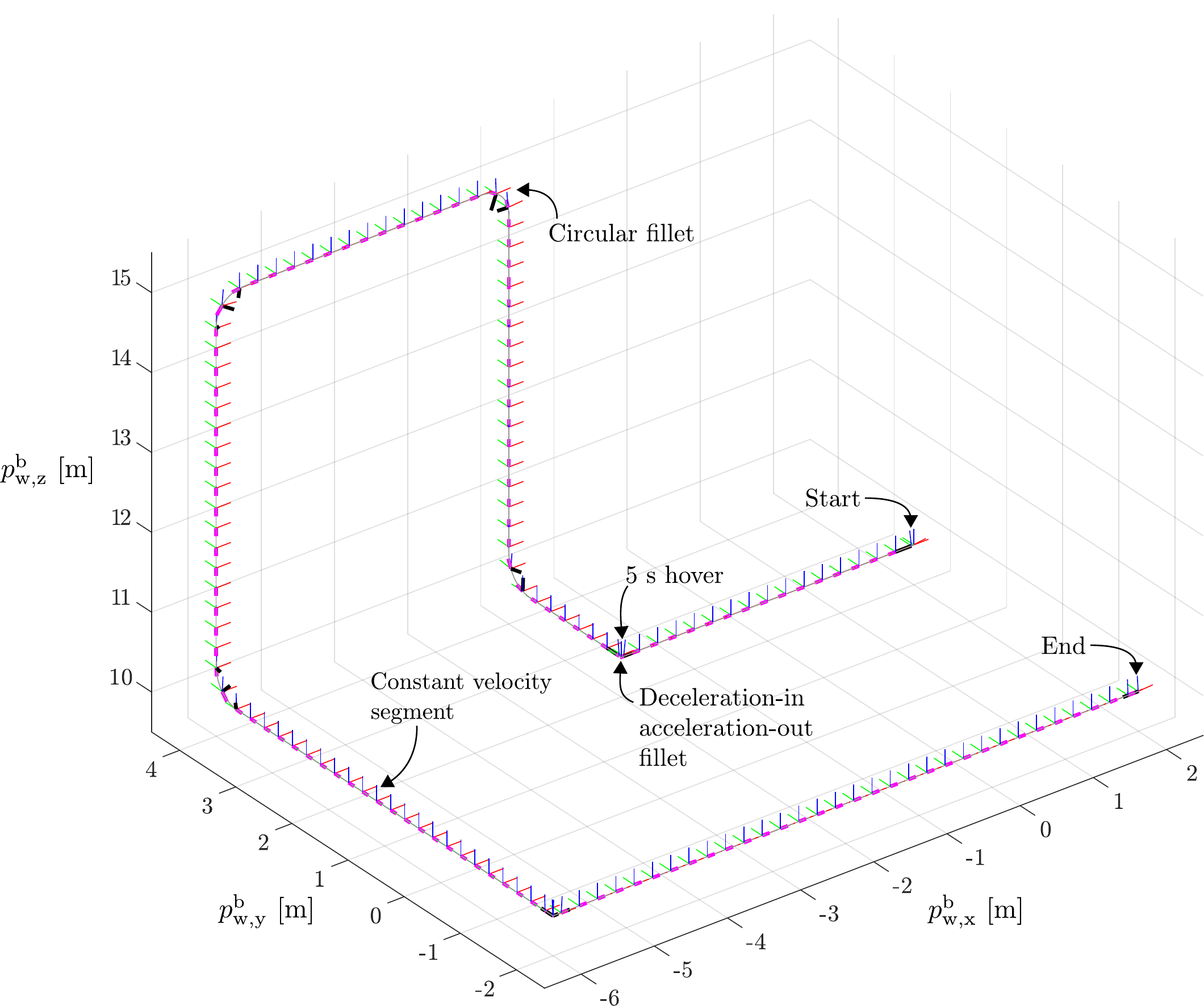}
  \caption{An example waypoint trajectory. Corresponding UAS body frame, magenta
    velocity and black acceleration are shown at 2 second time intervals. Body
    frame attitude is obtained via the flat output to state map
    \cite{Mellinger2011}.}
  \label{fig:waypoint_trajectory_example_full}
\end{figure}

A mission is comprised of a sequence of trajectory segments. As shown in
Figure~\ref{fig:trajectory_illustration}, there are two main types of
trajectories. The first, a transfer trajectory, interpolates all derivatives up
to snap smoothly between any two values in a fixed amount of time. The second, a
mission trajectory, is a set of constant velocity segments between user-defined
waypoints where transitions between different velocities are filleted by
transfer trajectories. Figure~\ref{fig:waypoint_trajectory_example_full} shows a
typical example of a generated waypoint mission, with the UAS body frame sampled
along its length. We also provide the ability to hover at any waypoint in the
mission trajectory, which occurs at the first and second to last waypoints in
Figure~\ref{fig:waypoint_trajectory_example_full}. It can be seen that the
trajectory consists mostly of constant velocity segments. Our approach ensures
that these segments between user-defined waypoints are straight, which is
desirable for most remote sensing applications.

\subsection{Trajectory Sequencer}

\begin{figure}
  \centering
  \begin{subfigure}[t]{.45\columnwidth}
    \centering
    \includegraphics[width=1\columnwidth]{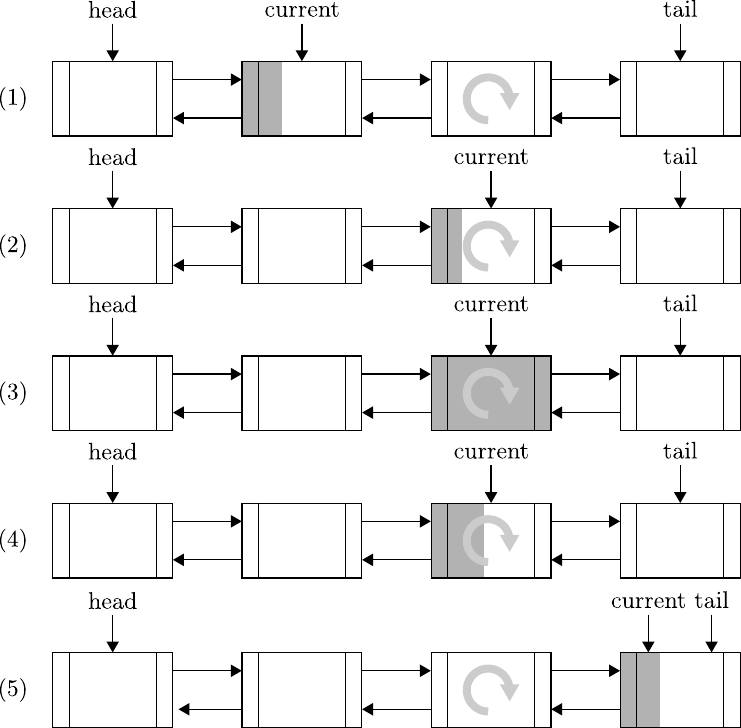}
    \caption{A trajectory sequence with a cyclic segment. From (1) to (2), the
      sequence is played forward until the cyclic segment is entered. From (2)
      to (4), an arbitrary time may pass as each time that the cyclic element is
      finished, it gets rewound. Therefore, \textit{current} never advances. In
      (5), the cyclic segment is aborted to move to the next element.}
    \label{fig:sequence_example_2}
  \end{subfigure}%
  \hfill%
  \begin{subfigure}[t]{.45\columnwidth}
    \centering
    \includegraphics[width=1\columnwidth]{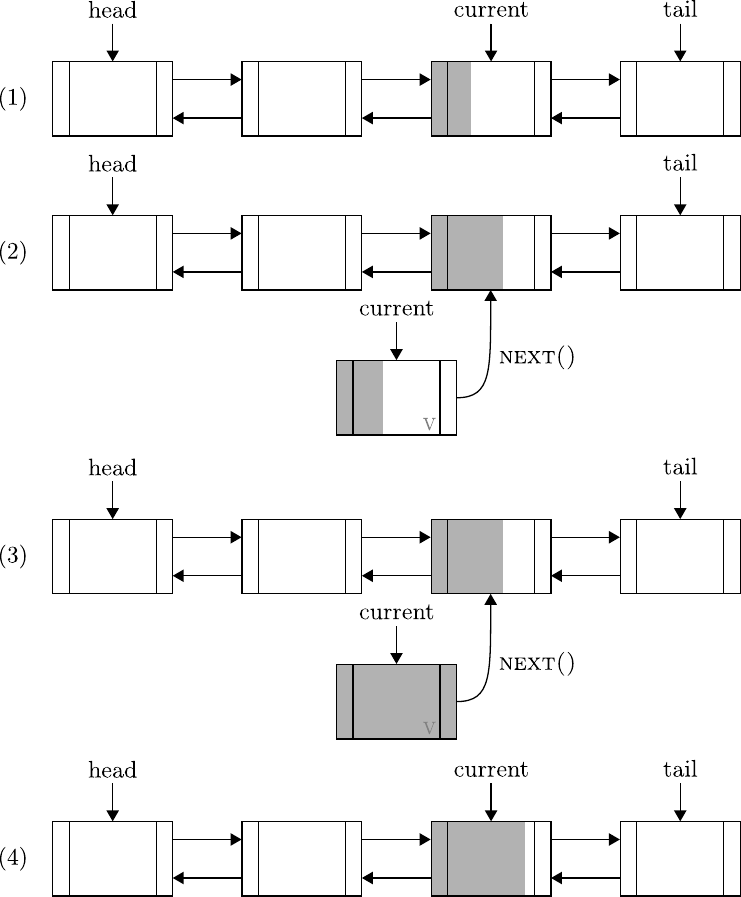}
    \caption{An example with a volatile element. Between (1) and (2), a volatile
      transfer trajectory is inserted. This pauses the original \textit{current}
      segment while the volatile segment is tracked. In (3), the volatile
      element is finished. The list transitions back to the original element
      while the volatile element is removed. List evaluation continues in (4).}
    \label{fig:sequence_example_3}
  \end{subfigure}%
  \caption{Trajectory sequencing examples using a doubly-linked list which moves
    the reference quadrotor state from one segment to the next in
    Figure~\ref{fig:trajectory_illustration}. Partial fill with a gray
    background is used to illustrate time along each segment.}
  \label{fig:sequence_examples}
\end{figure}

The trajectory sequencer aggregates individual trajectory segments into a
continuous sequence of trajectories.
This is implemented as a doubly-linked list where each element stores the
trajectory segment, a timestamp and a type which is either single or cyclic
where in the former case the trajectory is flown only once and in the latter it
is repeated until being aborted by other means such as a low battery charge
flag. As described in Figure~\ref{fig:sequence_example_3}, we additionally
provide a volatile element which enables executing one-time transfer
trajectories to correct for very large trajectory tracking errors.

\section{Control}
\label{sec:control}

\begin{figure}
  \centering
  \includegraphics[width=0.8\textwidth]{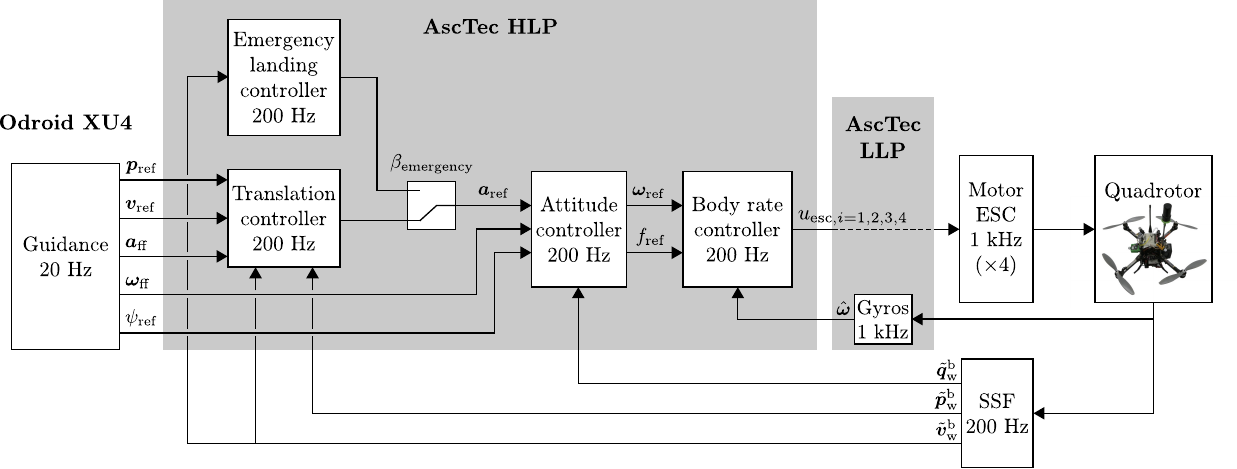}
  \caption{Cascaded control loop block diagram with the following nested loops:
    guidance, translation, attitude and body rates. Translation control
    generates a reference thrust vector, attitude control generates desired body
    rates to achieve it and body rate control generates the desired motor
    thrusts. Motor control is performed by AscTec ESCs in open loop.}
  \label{fig:cascaded_control_loop}
\end{figure}

We use a three-stage cascaded control architecture shown in
Figure~\ref{fig:cascaded_control_loop} to track the reference trajectories
output by the trajectory sequencer. The stages are, from outer- to inner-most
loop: translation, attitude and body rate controllers. A cascaded controller has
the advantage that inner loops provide faster disturbance rejection and reduce
the effect of nonlinearities \cite{Skogestad2005}, which is beneficial for
accurate tracking.  The controller is executed on the AscTec HLP as shown in
Figure~\ref{fig:hardware_architecture_quadrotor}.

The guidance subsystem forms the outermost loop. It is executed at 20 Hz and
outputs a reference position $\bm{p}_{\text{ref}}$, a reference velocity
$\bm{v}_{\text{ref}}$, a reference yaw $\psi_{\text{ref}}$ and optionally a
feed-forward acceleration $\bm{a}_{\text{ff}}$ and feed-forward body rates
$\bm{\omega}_{\text{ff}}$. All quantities are obtainable from our polynomial
trajectories via differential flatness theory \cite{Mellinger2011}.

Translation control forms the next inner loop and consists of a pre-filtered two
degree of freedom PID controller \cite{Skogestad2005} which outputs a desired
vehicle acceleration $\bm{a}_{\text{ref}}$. Since $\bm{a}_{\text{ref}}$ defines
a reference thrust vector, the translation controller may be thought of as a
thrust vector calculator. In order to prevent excessive tilt of the UAS, we
limit the thrust vector to a $20^\circ$ cone half-angle about the vertical
axis. Translation control loop bandwidth is between 1 and 2~rad/s. In the event
of a loss of state estimation due to communication or sensor failure, a standby
emergency landing controller can override the translation controller and land
the UAS by inducing a $-2$ m/s vertical velocity in an open-loop fashion similar
to \cite{Mueller2012,Lupashin2014}. This event can occur as a result of serial
communication cable failure, for example.

The attitude control forms the next inner loop and tracks $\bm{a}_{\text{ref}}$
and $\psi_{\text{ref}}$ by generating a body rate reference
$\bm{\omega}_{\text{ref}}$ and a collective thrust reference $T_{\text{ref}}$
for the body rate controller.  Our implementation of the attitude controller is
based on the globally asymptotically stable and robust to measurement noise
quaternion-based controllers introduced in
\cite{Brescianini2013,Faessler2015}. The attitude control loop bandwidth is
between 5 and 10~rad/s.

The body rate controller forms the inner-most loop of the cascaded flight
control system. It tracks $\bm{\omega}_{\text{ref}}$ by computing reference body
torques, following the feedback-linearizing control scheme presented in
\cite{Faessler2016}. These along with $T_{\text{ref}}$ are then converted into
individual reference motor thrusts $f_{\text{ref}}$ that are mapped to propeller
speeds via a full quadratic motor calibration map. Our controller uses a thrust
saturation scheme based on \cite{Faessler2016} which prioritizes roll and pitch
torques, which are the most important ones for stability. The body rate control
loop bandwidth is approximately 40~rad/s.

\section{Autonomy Engine}
\label{sec:autonomy_engine}

The autonomy engine subsystem implements the logic for long-duration fully
autonomous operation with repeated data acquisition flights.

\subsection{Architecture Overview}

\begin{figure}
  \centering
  \includegraphics[width=0.5\textwidth]{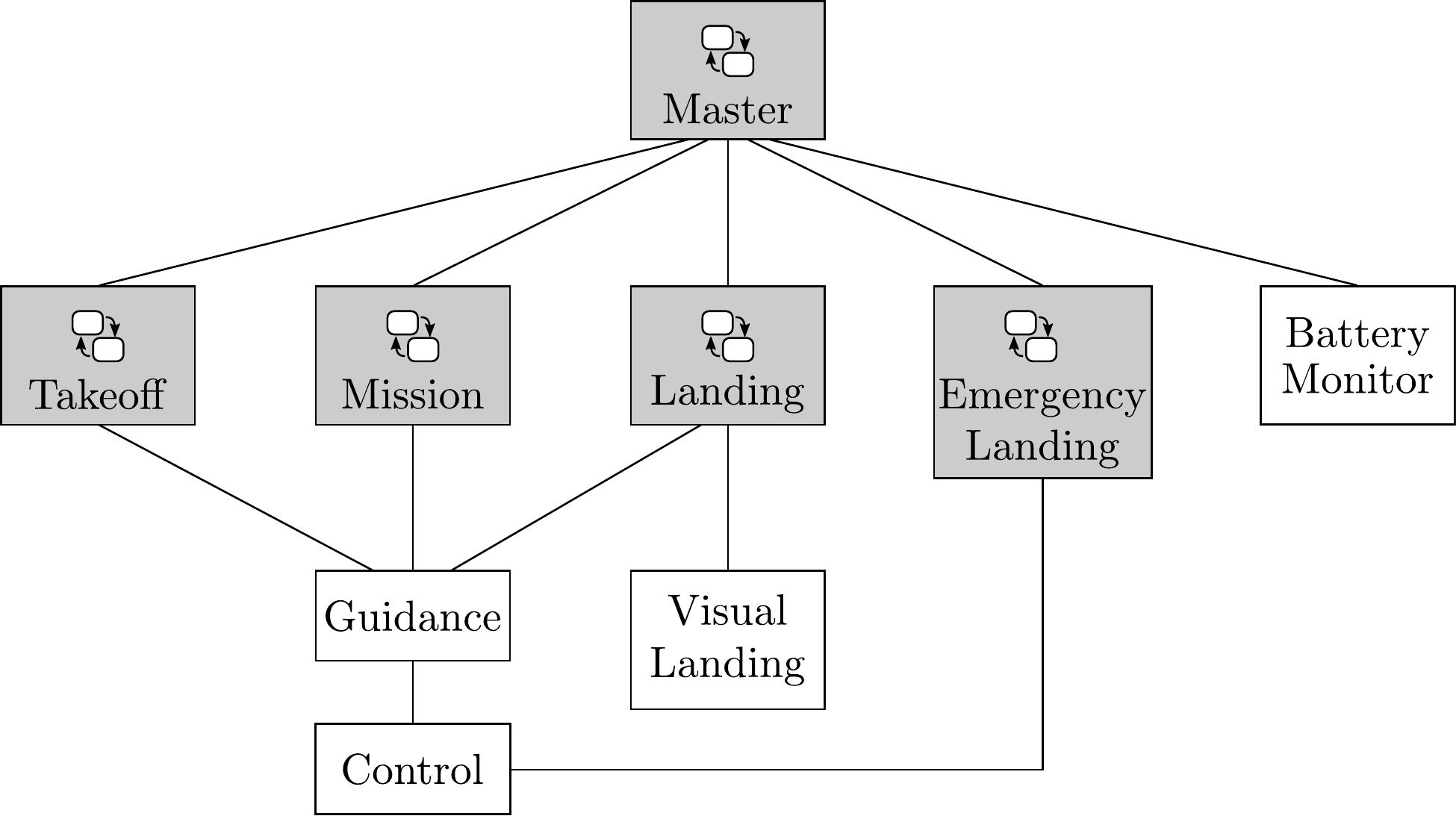}
  \caption{Hierarchy of master state machine and autopilots (gray) and other
    subsystems. The autopilots execute the logic associated with the
    corresponding mission phase and their execution is synchronized by the
    master to perform a specific mission profile.}
  \label{fig:sm_relationship}
\end{figure}

Figure~\ref{fig:mission_scheme} illustrates the overall concept of operation for
repeated data acquisition. Each mission phase is implemented as a stand-alone
state machine called an \textit{autopilot} that executes the logic associated
with the phase whenever it is activated via a Robot Operating System (ROS)
action interface. To coordinate calls to each autopilot, an additional
overarching logic is implemented as a high-level master state machine. The
hierarchy of state machines and their interface to other major subsystems is
shown in Figure~\ref{fig:sm_relationship}. All state machines are executed in
parallel threads at a frequency of 20~Hz.

\subsection{Master State Machine}

\begin{figure}%
  \centering%
  \includegraphics[width=0.5\columnwidth]{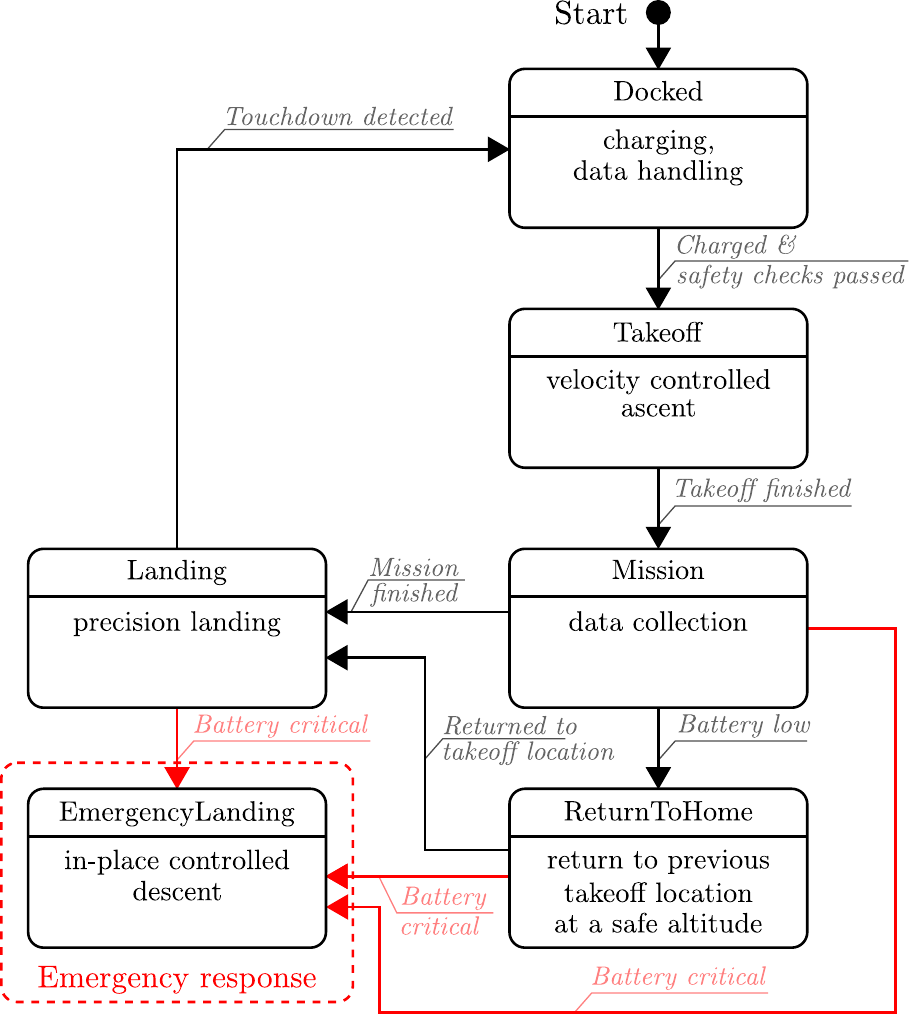}%
  \caption{Master state machine diagram. The master state machine parses mission
    events (e.g. battery status and touchdown detection) and coordinates calls
    to the phase-specific autopilots, but performs no computations itself.}%
  \label{fig:master_state_machine}%
\end{figure}

The master state machine coordinates calls to the phase-specific autopilots and
is illustrated in Figure~\ref{fig:master_state_machine}. In contrast to the
computation-heavy autopilots, the master performs no computations itself.  Its
sole responsibility is to initiate event-based state transitions. In the
takeoff, mission, landing and emergency landing states the master state machine
activates the appropriate autopilot and waits for it to complete. If necessary,
the master can also abort each autopilot in order to execute robust behaviors
for cases like low battery charge in flight or abnormal motor performance before
takeoff. For example, if battery charge anomalously drops to critical levels,
the mission and landing actions can be aborted mid-execution to perform an
emergency landing action.

\subsection{Takeoff Autopilot}

\begin{figure}
  \centering
  \begin{subfigure}[b]{0.48\columnwidth}
    \includegraphics[width=1\columnwidth]{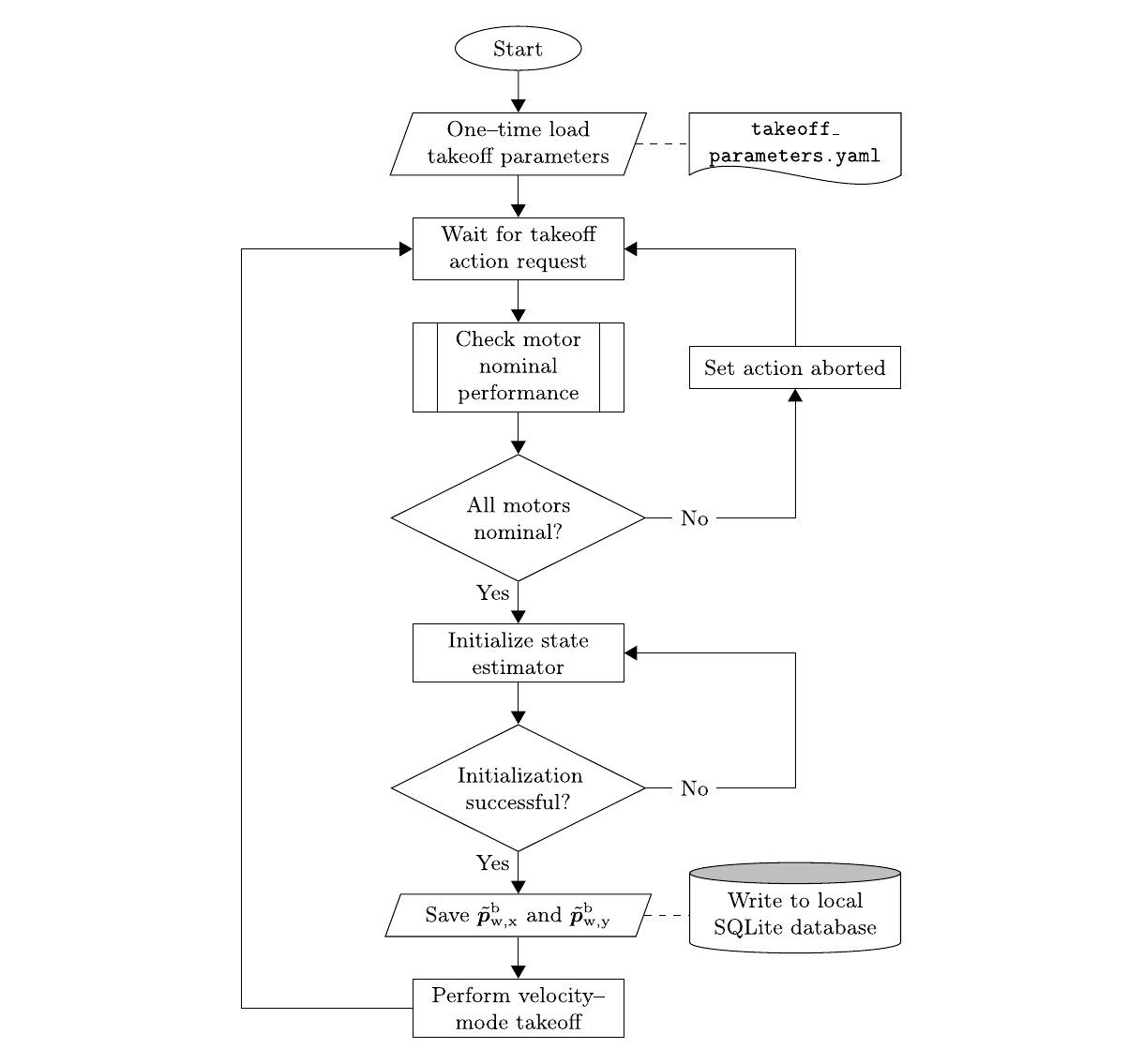}
    \caption{Takeoff autopilot. This takes the UAS from a motors-off
      state on the charging pad to a hover at a target takeoff altitude.}
    \label{fig:flowchart_takeoff}
  \end{subfigure}%
  \hfill%
  \begin{subfigure}[b]{0.48\columnwidth}
    \includegraphics[width=1\columnwidth]{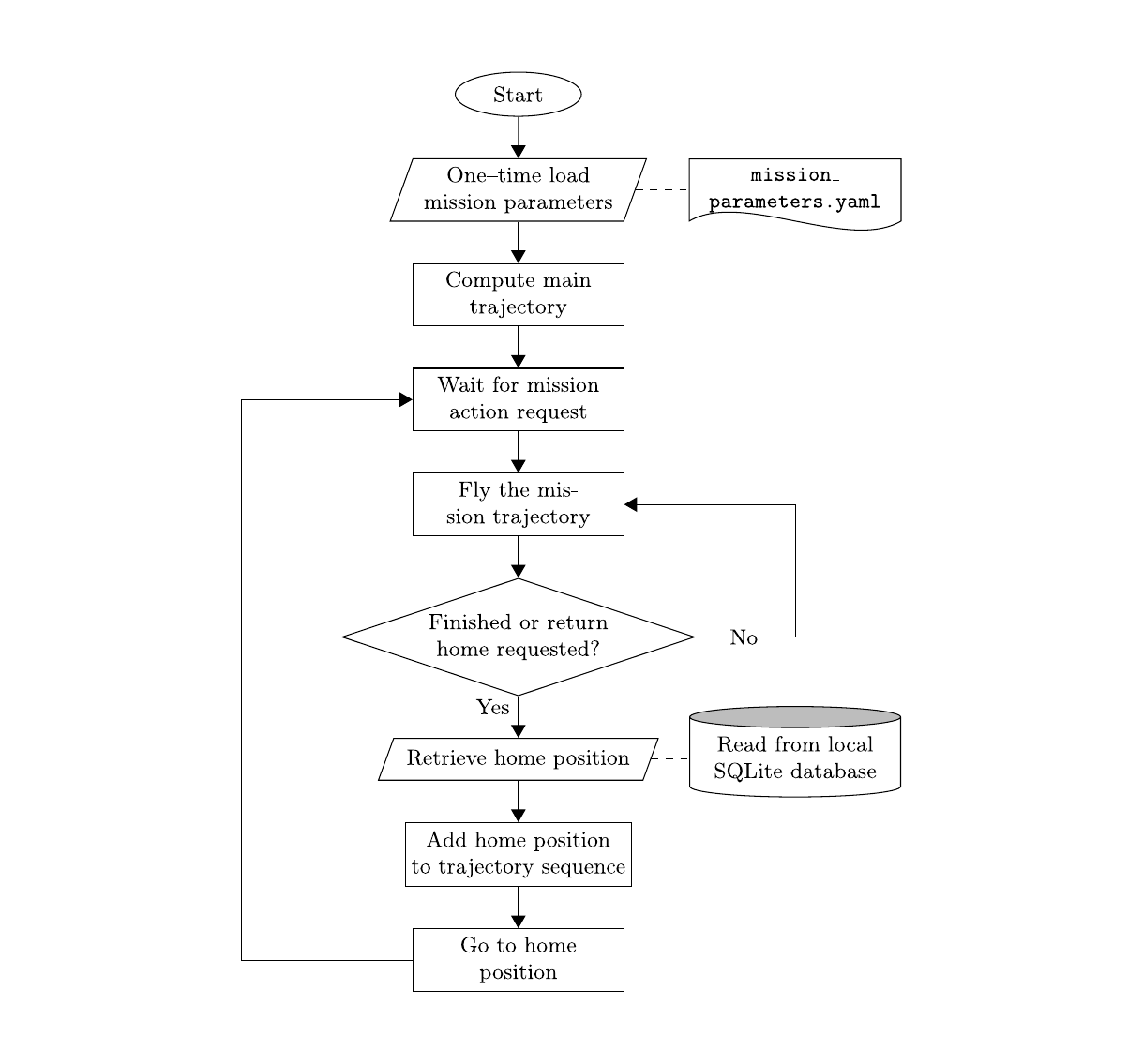}
    \caption{Mission autopilot. This executes the actual data acquisition
      mission by interfacing with the guidance subsystem
      (Section~\ref{sec:guidance}).}
    \label{fig:flowchart_mission}
  \end{subfigure}

  \begin{subfigure}[b]{0.48\columnwidth}
    \includegraphics[width=1\columnwidth]{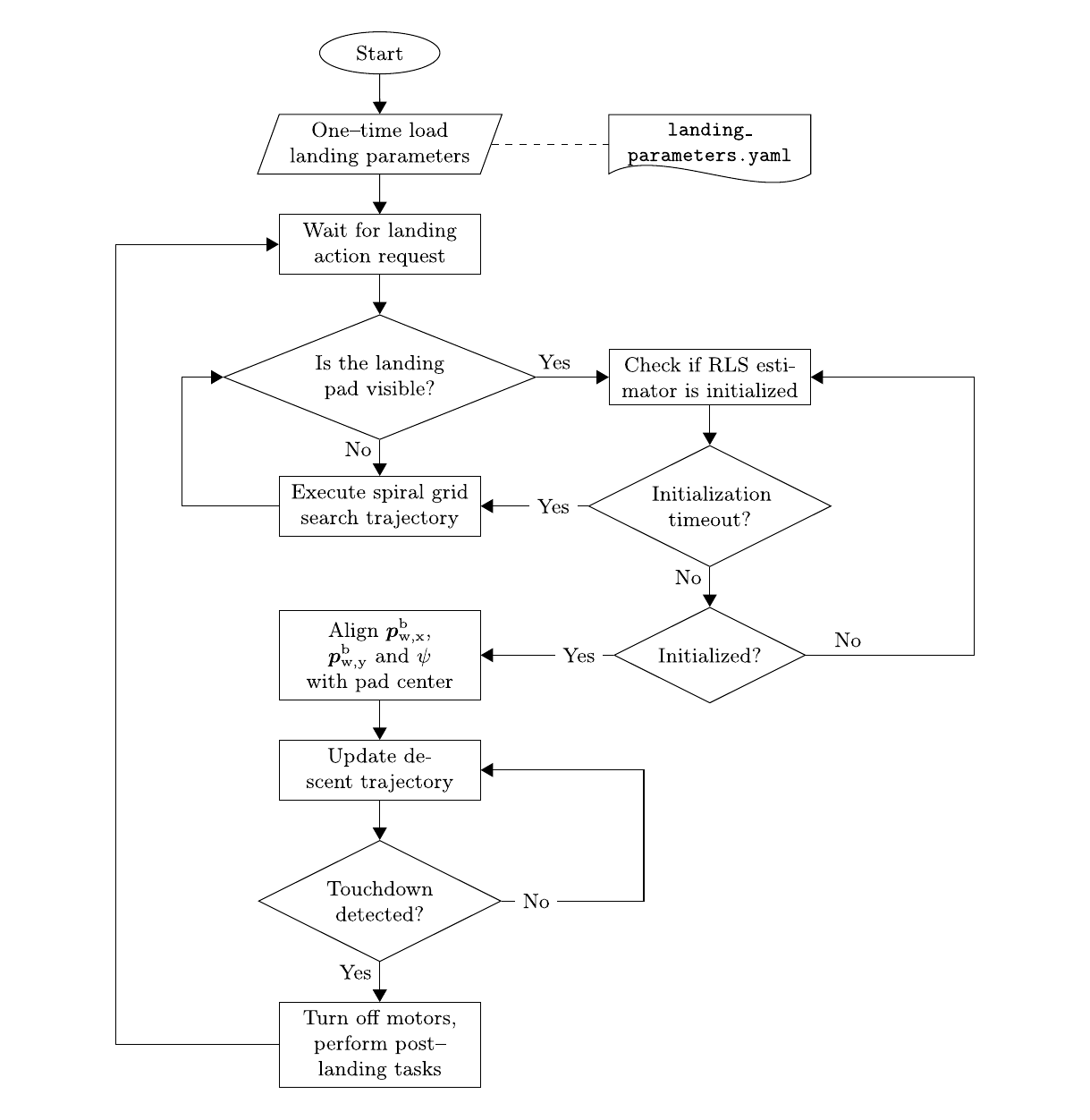}
    \caption{Landing autopilot. This takes the UAS from a hovering state in the
      vicinity of the landing pad to a motors-off state on the charging
      surface.}
    \label{fig:flowchart_landing}
  \end{subfigure}%
  \hfill%
  \begin{subfigure}[b]{0.48\columnwidth}
    \includegraphics[width=1\columnwidth]{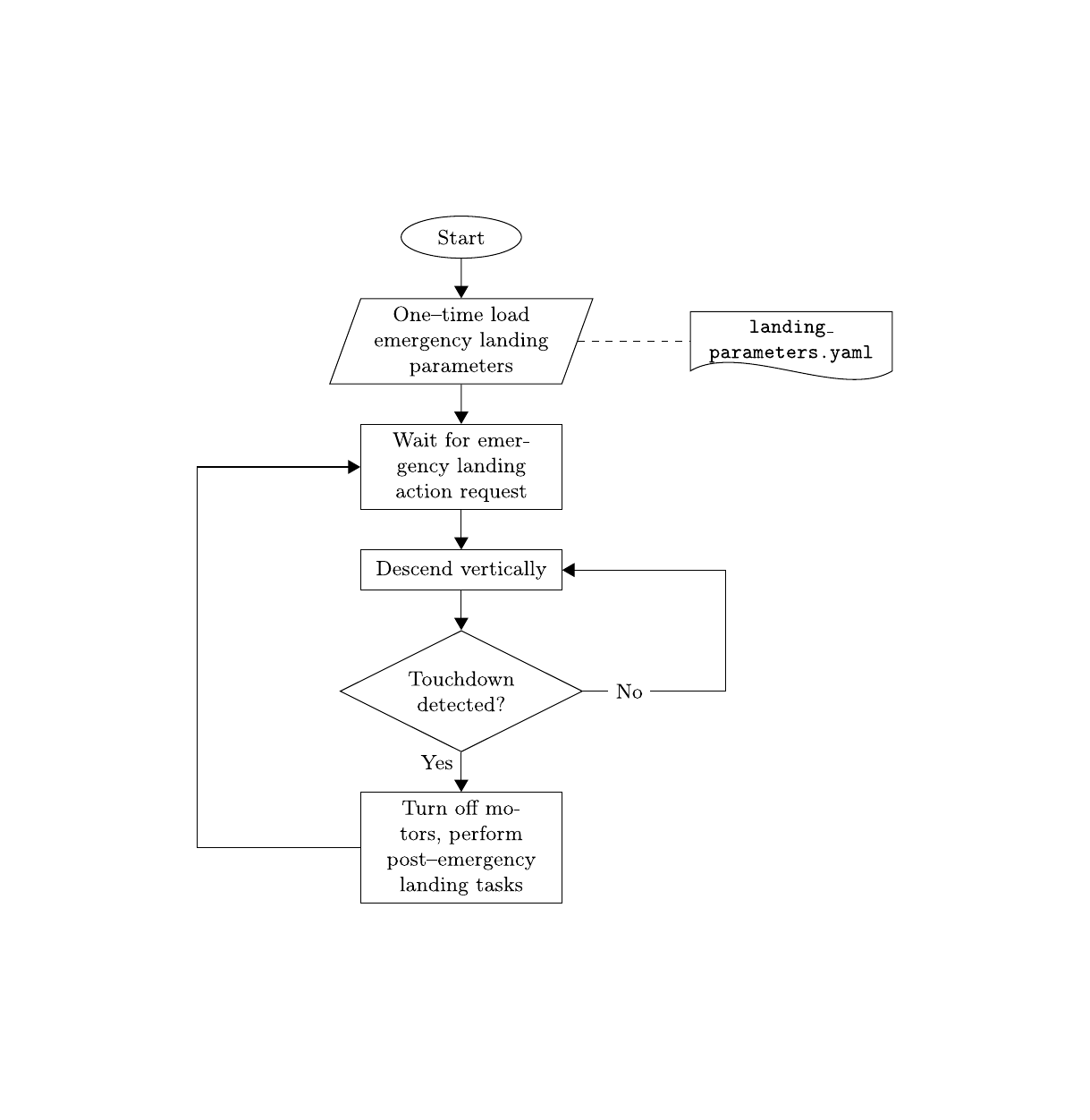}
    \caption{Emergency landing autopilot. This brings the UAS to a soft
      touchdown at its current location when a critically low battery event
      occurs.}
    \label{fig:flowchart_emergency}
  \end{subfigure}

  \caption{Autopilot logic in flowchart representation.}
  \label{fig:flowcharts}
\end{figure}

The takeoff autopilot takes the UAS from a motors-off state on the charging pad
to a hover at a target takeoff altitude. The procedure is illustrated in
Figure~\ref{fig:flowchart_takeoff}. In summary, after successfully validating
motor nominal performance, re-initializing the state estimator after the
prolonged charging phase and memorizing the current horizontal location in
permanent memory, the takeoff autopilot commands a velocity-controlled takeoff
that takes it from the initial position on the charging pad to a target
altitude.

Two parts of the takeoff procedure are of particular interest. The first is a
motor nominal performance check which occurs prior to takeoff. Motors are spun
at a low RPM, which is measured via zero crossing detection, and are verified to
rotate within 400 RPM of the nominal value. Ten attempts to pass this check are
allowed before the takeoff is aborted. Our test flights had instances of this
procedure saving the UAS e.g. when an I$^2$C cable to one of the motors
dislodged. The next important part of the takeoff is saving the current location
to a database in permanent memory. This gives the UAS knowledge of where to
return to after the mission and also prevents this information from being lost
in case a software reset is required.

\subsection{Mission Autopilot}

The mission autopilot is responsible for executing the actual data acquisition
mission as defined by the user in the form of individual waypoints and hover
times. The algorithm is illustrated in Figure~\ref{fig:flowchart_mission}. The
mission trajectory is performed either once or is re-flown continuously until an
event such as low battery charge requests the UAS to return to the charging pad.

\subsection{Landing Autopilot}
\label{subsec:landing_autopilot}

The landing autopilot safely takes the UAS from a hovering state in the vicinity
of the landing pad to a motors-off state on the charging surface. The procedure
is illustrated in Figure~\ref{fig:flowchart_landing}. The first action is to
check if the landing pad is visible in the downfacing navigation camera image,
i.e. if any AprilTag in the bundle is detected. If not, the UAS executes the
spiral grid search trajectory until the landing bundle becomes visible. The
vision-based landing navigation algorithm
(Figure~\ref{fig:visual_navigation_block_diagram}) then becomes activated. The
estimated yaw is used to align the vehicle attitude such that the camera points
towards the visual markers, while the landing position estimate is used to align
the vehicle's lateral position over the center of the charging pad. The vehicle
then performs a constant velocity descent until touchdown is detected based on
0.3 m height and 0.1 m/s velocity thresholds:
\begin{equation}
  \mathrm{touchdown} =
  (\tilde{p}_{\text{w,z}}^{\text{b}}-\tilde{p}_{\text{w,z}}^{\text{l}})<0.3
  \,\,\,\wedge\,\,\,
  |\tilde v_{\text{w,z}}^{\text{b}}|<0.1.
\end{equation}

\subsection{Emergency Lander}

The emergency lander brings the UAS to a soft touchdown at its current location
and is triggered in response to a critically low battery voltage. Note that this
is a different behavior from the emergency landing controller in
Figure~\ref{fig:cascaded_control_loop}, which handles state estimation
failure. The emergency lander logic is illustrated in
Figure~\ref{fig:flowchart_emergency}. The algorithm descends the UAS with a mild
vertical velocity of 0.3 m/s until touchdown is detected based on a 0.1 m/s
velocity threshold:
\begin{equation}
  \mathrm{touchdown} =\|\tilde{\bm{v}}_{\text{w}}^{\text{b}}\|_2<0.1.
\end{equation}

\section{Experimental Results}
\label{sec:results}

The system was tested in four configurations of increasing complexity. These
were software-in-the-loop (SITL) simulation \cite{rotors:2016}, indoor tethered
flights with VICON-based state estimation, outdoor tethered flights with
GPS-based state estimation and outdoor free flights. SITL simulation allows for
rapid testing and debugging by removing most logistic concerns and by making it
easier to identify algorithmic errors due to the absence of real-world
noise. Indoor test flights under VICON allow to refine the guidance, control and
autonomy subsystems independently of state estimation. Finally, outdoor flights
validate that the system works in its real-world environment.

\subsection{Vision-based Landing Accuracy}

To design an AprilTag marker bundle for localizing the landing pad as explained
in Section~\ref{subsec:visual_landing_navigation}, we first analyzed the
detection accuracy in a stationary setting of the AprilTag 2 algorithm for
single tags at various observation distances using the downfacing camera that is
deployed on our UAS (see Figure~\ref{fig:quad_labeled}).
\begin{figure}%
  \centering
  \includegraphics[width=0.5\columnwidth]{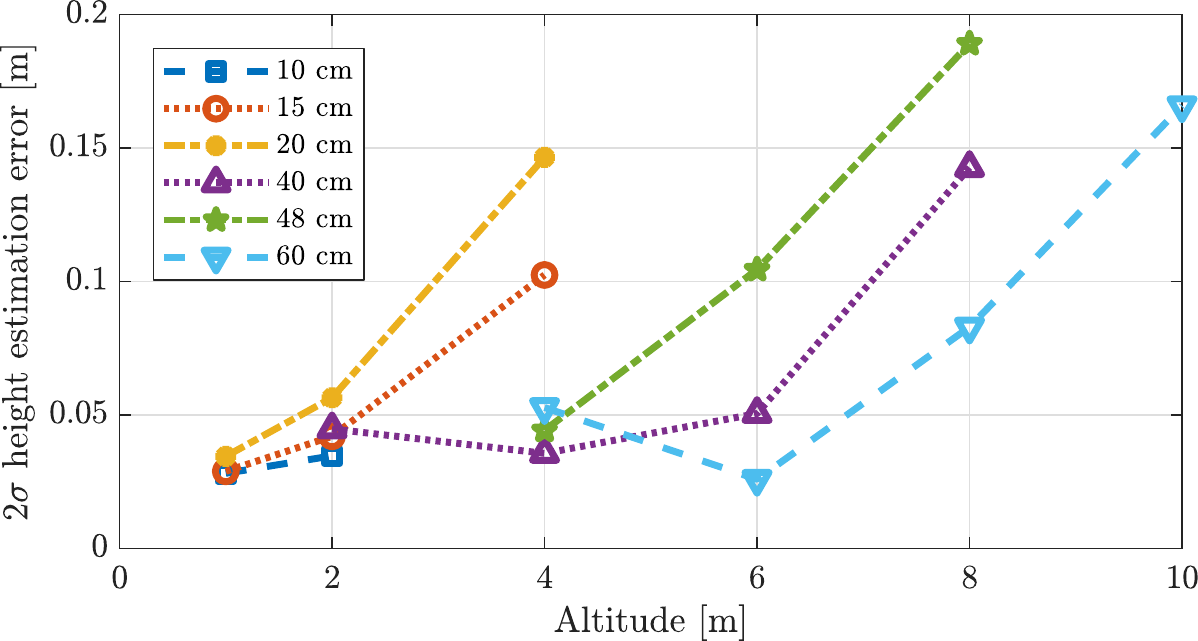}%
  \caption{Distance estimation error of AprilTag 2 detector for various tag
    sizes and heights. For a given tag size, accuracy decreases as distance
    increases. For large enough distances, smaller tags cease to be detected.}%
  \label{fig:april_tag_height_error}%
\end{figure}
\begin{figure}%
	\centering
	\includegraphics[width=0.9\columnwidth]{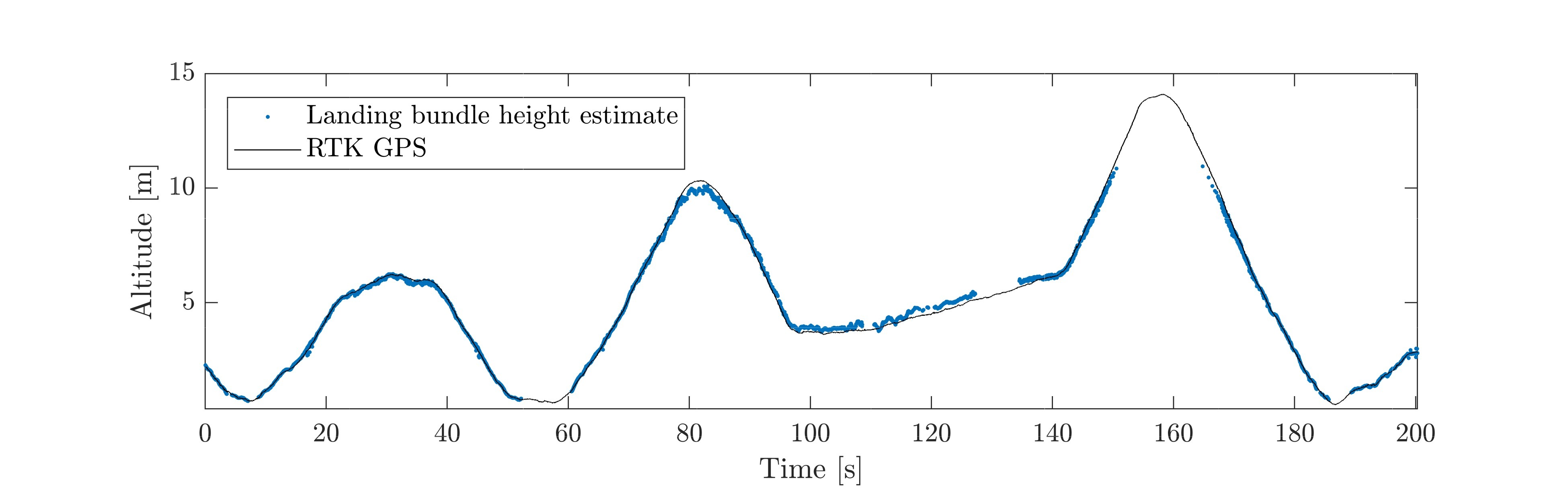}%
	\caption{Comparison of landing bundle height estimate with ground truth
          altitude during a test flight of the UAS above the landing
          station. Ground truth is provided by the RTK-GPS altitude reading.
        }%
	\label{fig:landing_target_height_accuracy}%
\end{figure}
\begin{figure}%
	\centering
	\includegraphics[width=0.8\columnwidth]{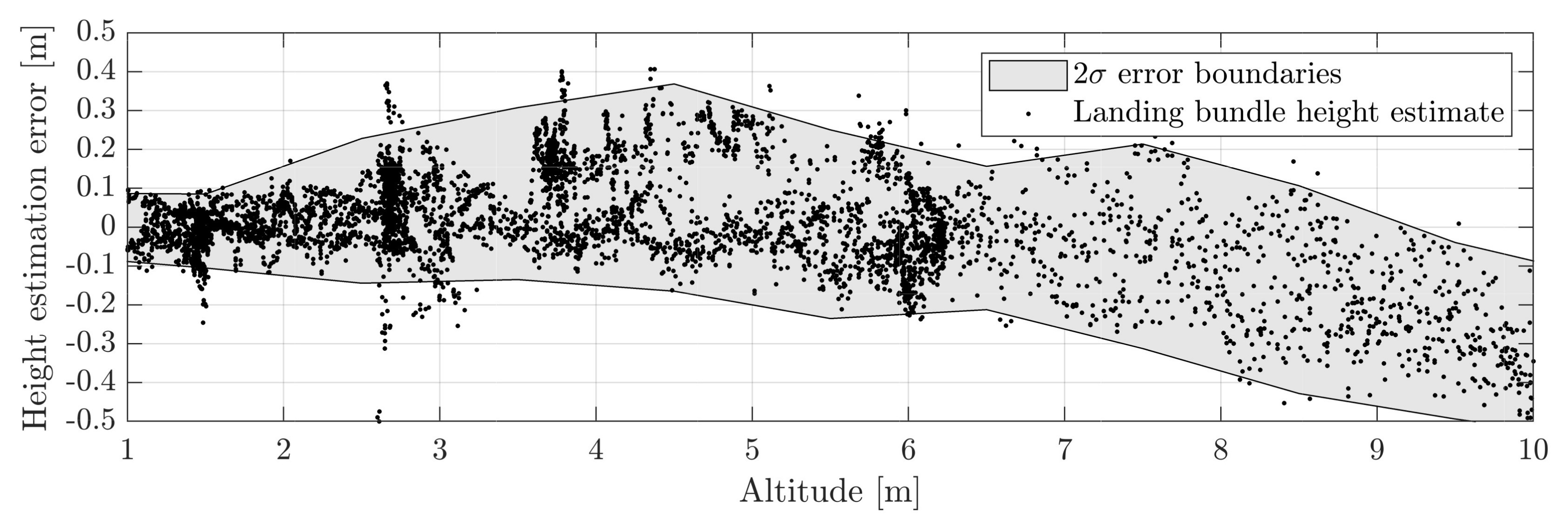}%
	\caption{Landing bundle height error as a function of flight
          altitude. The error is approximately 0.32~m at the approach height of
          4~m and decreases to below 0.1~m in the final stage of descent.}%
	\label{fig:landing_target_height_accuracy_error}%
\end{figure}

Figure~\ref{fig:april_tag_height_error} shows the standard deviation of the
distance estimate for various tag sizes and distances.  For a given tag size,
the estimation accuracy decreases with increasing distance.  Large tag sizes
(48~cm, 60~cm) are not detected at distances below 2~m, while small tag sizes
(10~cm, 15~cm, 20~cm) cease to be detected above 4~m.  For an initial approach
height of 4~m, we selected a 48~cm tag for an expected $2\sigma$ (i.e. 2
standard deviation) height estimation error of 4.5~cm and bundled it with three
15~cm tags for increased precision at lower altitudes, resulting in the bundle
depicted in Figure~\ref{fig:landing_pad_with_QR} and the bottom left of
Figure~\ref{fig:bundles_test}.

Figure~\ref{fig:landing_target_height_accuracy} illustrates an experiment where
the quadrotor was flown at various heights above the landing marker bundle shown
in Figure~\ref{fig:landing_pad_with_QR}. The statistics for the error between
the height estimate using the AprilTag detector and the ground truth from
RTK-GPS are plotted in Figure~\ref{fig:landing_target_height_accuracy_error}.
At the approach height of 4~m the 2$\sigma$ height detection error is
0.32~m. The error decreases to below 0.1~m in the final stage of the descent. We
note that this estimation error variance is larger than for the individual tags
in Figure~\ref{fig:april_tag_height_error}. The increased error is most likely
caused by image blur and latency effects due to the motion of the vehicle and a
lever arm effect that amplifies position errors at the desired landing location
(the charging pad center) as a result of angular errors in the AprilTag
detection. Section~\ref{subsubsec:optimal_bundle_layout} provides a remedy to
the lever arm effect.

\begin{figure}%
  \centering
  \includegraphics[width=0.9\columnwidth]{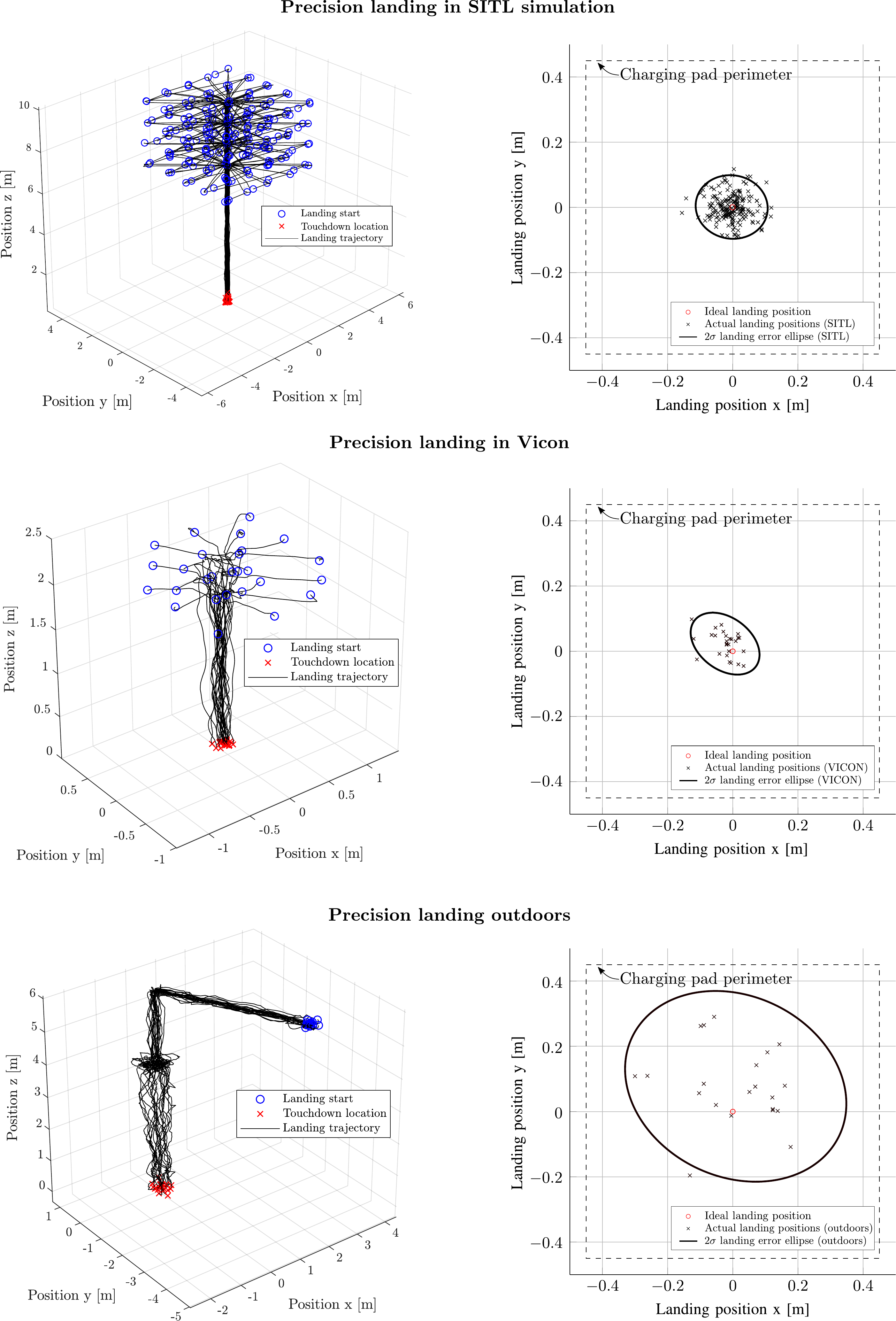}
  \caption{Accuracy of touch-down position in SITL simulation with severe wind
    (top row), indoors in ideal conditions (middle row) and outdoors with light
    wind (bottom row). Ground truth is provided indoors by VICON and outdoors by
    RTK-GPS. SITL simulation uses a noise-free state.}
  \label{fig:landing_ellipses}%
\end{figure}

Figure~\ref{fig:landing_ellipses} illustrates the accuracy of autonomous landing
in three environments of increasing complexity. First, 200 landings were tested
in SITL simulation with a random wind force averaging 30 km/h modeled as a
random walk for the wind direction. The $2\sigma$ landing error is 0.09~m (major
half-axis) and 0.08~m (minor half-axis). Because simulation uses the
ground-truth UAS pose for control, we may conclude that this is roughly the
landing error that our control and landing navigation algorithms are able to
maintain in severe wind conditions. Next, the UAS performed 27 landings indoors
with VICON state estimation in near-optimal conditions. The resulting $2\sigma$
landing error ellipse aggregates all detection and control errors. The lateral
$2\sigma$ error is 0.11~m (major half-axis) and the longitudinal $2\sigma$ error
is 0.08~m (minor half-axis). We see that the accuracy is similar to simulation,
which indicates that real-world uncertainty sources like model inaccuracy and
data signal latency degrade landing precision by a similar amount as severe wind
conditions with an otherwise ideal system. Finally, 21 landings were performed
outdoors using RTK-GPS for ground truth. The lateral $2\sigma$ error is 0.37~m
(major half-axis) and the longitudinal $2\sigma$ error is 0.28~m (minor
half-axis). While this is well within the charging surface dimensions, the
increased landing error is representative of the less accurate outdoor vehicle
pose estimate, camera exposure adjustment effects due to light conditions, more
complex aerodynamic effects from wind than those modeled in simulation,
etc. Because we required the downfacing camera to collect more pad pose
measurements in order to initialize the RLS filter from
Section~\ref{subsec:visual_landing_navigation}, the outdoor landing trajectories
have a visible hover period at the preset 4~m approach altitude in the bottom
left of Figure~\ref{fig:landing_ellipses}.

\subsubsection{Optimizing the AprilTag Bundle Layout}
\label{subsubsec:optimal_bundle_layout}

\begin{figure}
  \centering
  \includegraphics[width=0.8\columnwidth]{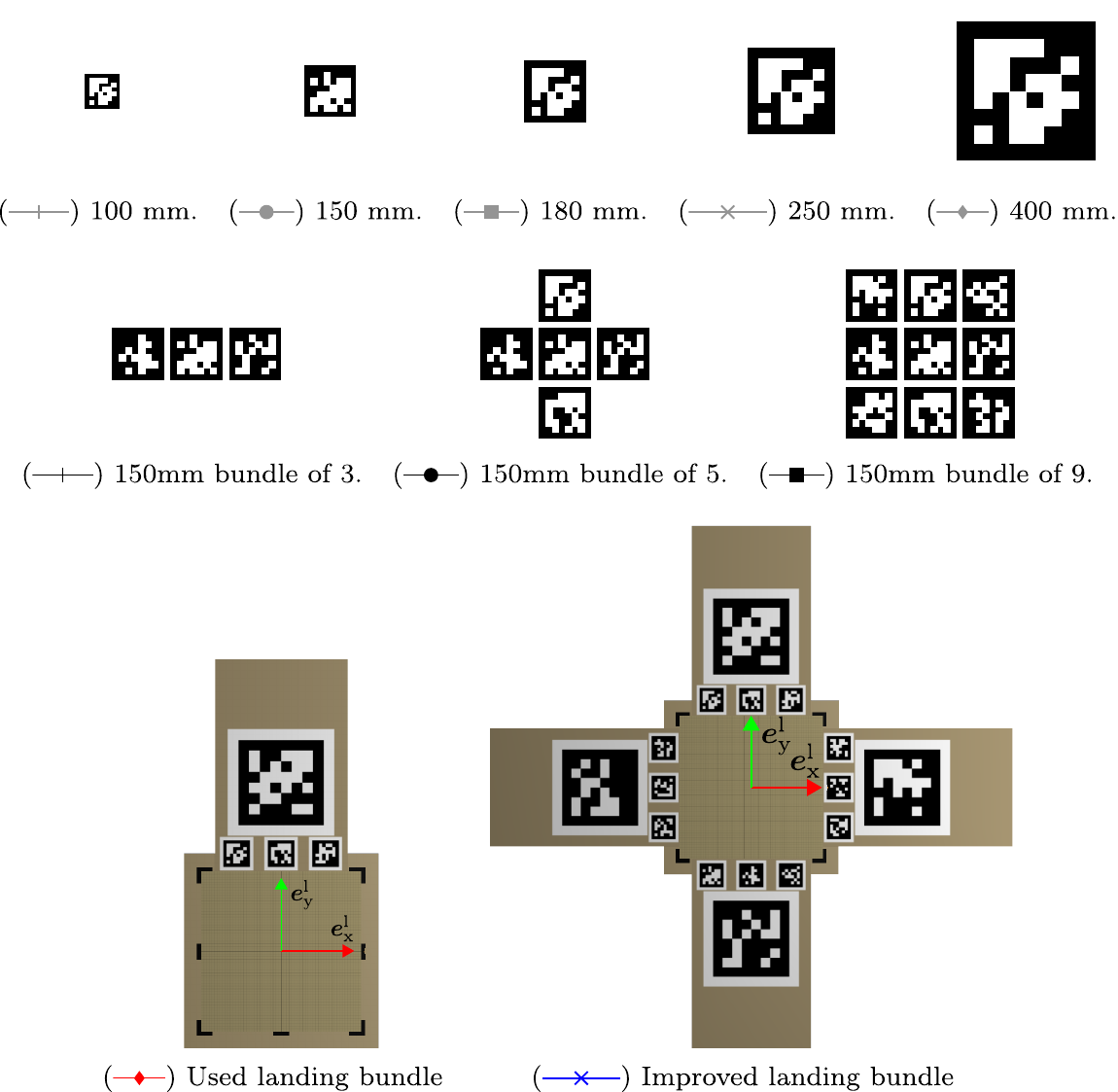}
  \caption{Bundles for simulated measurement noise data collection. The goal is
    to see how individual tag size and bundle geometry affect detection
    accuracy. The last two landing bundles consist of~480 mm and 150~mm tags.}
  \label{fig:bundles_test}
\end{figure}

\begin{figure}
  \centering
  \includegraphics[width=0.8\columnwidth]{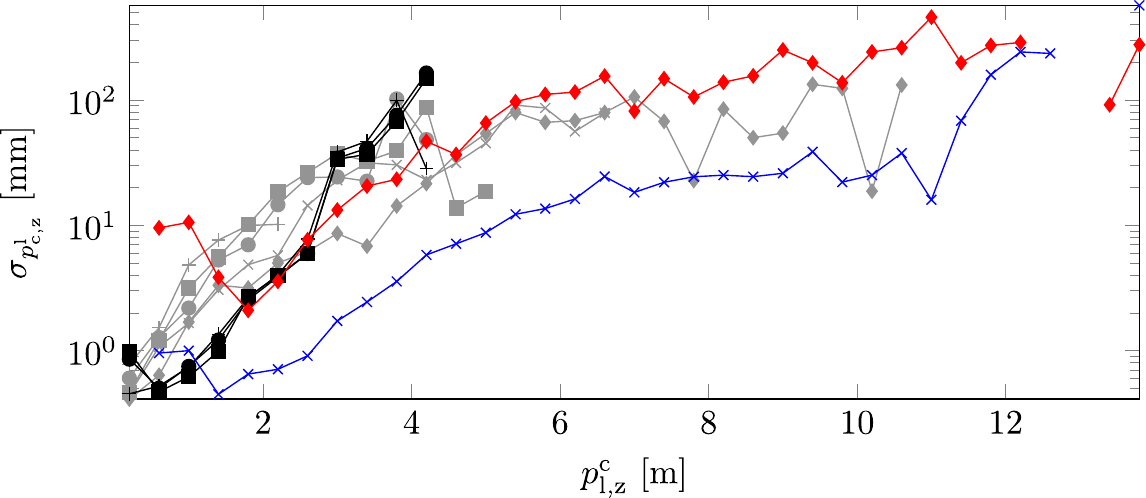}
  \caption{Height detection noise variance for the simulated tag bundles. Each
    line corresponds to a tag in Figure~\ref{fig:bundles_test}. Larger tags
    decrease measurement noise and can be detected from afar. By bundling these
    with smaller tags, the bundle can also be detected from small distances. By
    distributing the tags symmetrically like for the improved landing bundle
    (blue curve), inaccuracies due to a lever arm effect are avoided.}
  \label{fig:at2_noise_analysis}
\end{figure}

We carried out a simulated set of experiments in Gazebo \cite{Koenig} to search
for an AprilTag bundle geometry which minimizes the landing pad position
measurement error. While many factors affect the AprilTag position measurement
error, such as the camera-tag off-axis angle \cite{olson2011tags,wang2016iros},
our investigation considers only the camera-tag distance while keeping the
off-axis angle zero. This choice is appropriate for our use case, since the
dominant variable that changes during the landing phase is the quadrotor height
and the camera is facing approximately straight down at the AprilTag
bundle. Furthermore, the simulation environment allows to isolate the effect of
bundle geometry since other variables (e.g. illumination, lens focus, etc.)  are
easily controlled for. Thus, bundle geometry and camera-tag distance are the
only independent variables. Our simulation uses the downfacing camera parameters
from Section~\ref{subsec:visual_landing_navigation}, i.e. a 752$\times$480~px
resolution and the same calibration matrix $K\in\reals^{3\times 3}$.

Figure~\ref{fig:bundles_test} shows the tag bundle geometries that were
tested. Note that the first two rows of bundles in Figure~\ref{fig:bundles_test}
are impractical to land on since they would obscure the charging surface. The
bundles in the third row are therefore placed around the charging area.

Figure~\ref{fig:at2_noise_analysis} shows the output of our simulation for the
height detection noise variance. We observe that a larger tag decreases
measurement noise (as seen for the top row single-tag bundles of
Figure~\ref{fig:bundles_test}). Importantly, the detection distance of larger
tags is larger because they make a bigger pixel footprint in the
image. Measurement noise for bundles in row 2 of Figure~\ref{fig:bundles_test}
is smaller than for the single-tag bundles, however the 5- and 9-tag bundles do
not appear to reduce noise with respect to the 3-tag bundle. By using both large
and small tags, the landing bundle we used in real-world experiments is able to
be detected from large distances of up to 14~m as well as from small distances
prior to touchdown. By offsetting the tags to one side of the charging station,
however, the detection accuracy is reduced by the aforementioned lever arm
effect. An improved bundle, with tags distributed symmetrically about the target
landing location, shows better detection performance. However, it is a less
practical solution as it takes up a lot more space.

\subsection{Autonomy Engine}

\begin{figure}
  \centering
  \includegraphics[width=1\columnwidth]{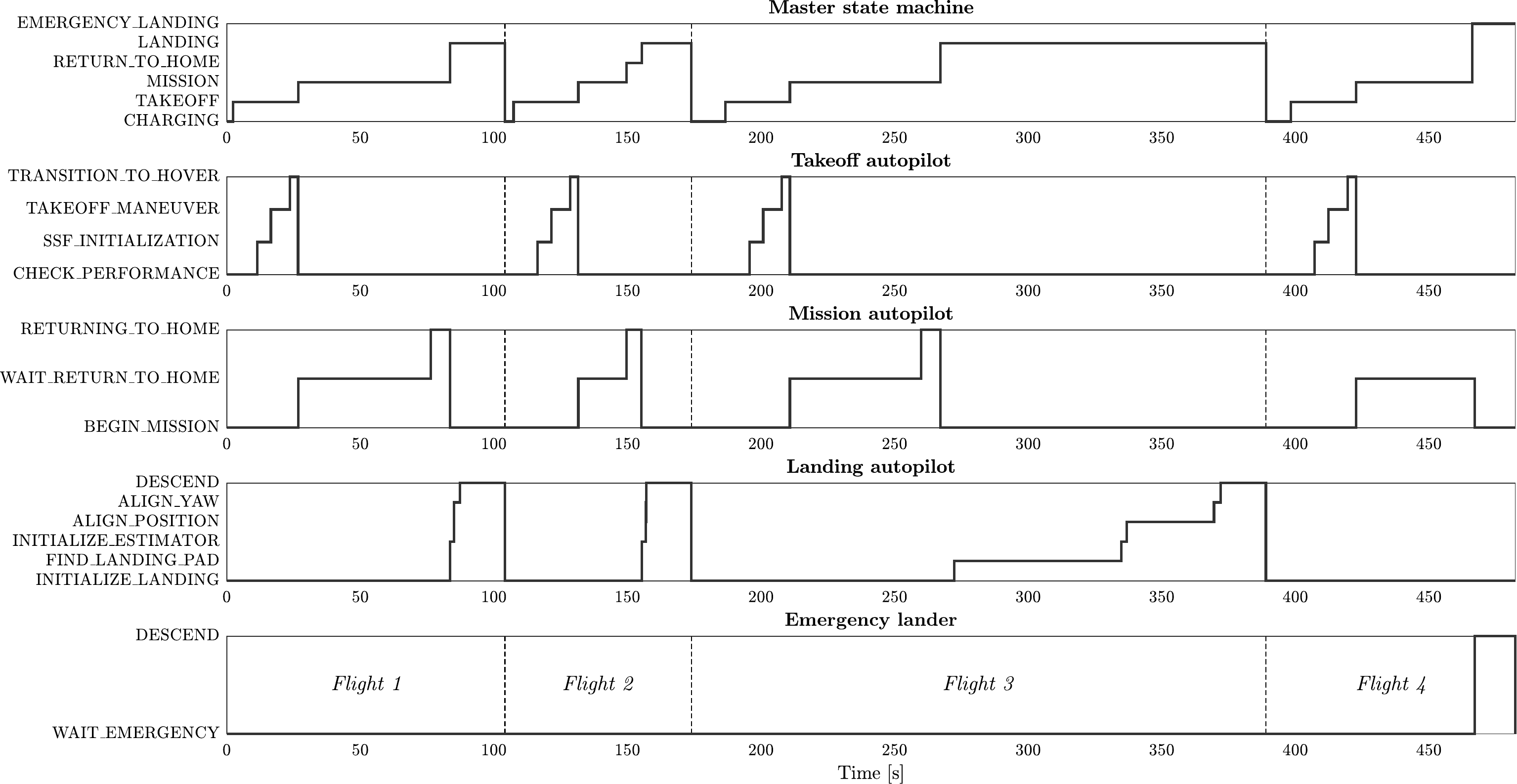}
  \caption{Autonomy engine states over four flights that are representative of
    the autonomy engine's dominant modes of operation. Dashed lines delimit
    individual flights (including on-pad time) and the flights are numbered in
    the bottom subplot. Flight (1) shows a nominal cycle, (2) shows an aborted
    mission due to a low battery, (3) shows a case where the landing pad is
    invisible upon returning home and (4) shows an emergency landing due to a
    critically low battery.}
  \label{fig:autonomy_states}
\end{figure}

Figure~\ref{fig:autonomy_states} shows the simultaneous operation of the master
state machine and the four autopilots for takeoff, mission, landing and
emergency landing in four simulated modes of operation. The first nominal flight
is the one illustrated in Figure~\ref{fig:mission_scheme}. In the second flight,
a low battery charge aborts the mission and the UAS returns to the charging
pad. In the third flight, the UAS spends some time executing the spiral grid
search trajectory because the landing marker bundle is initially not visible. In
the fourth and final flight, a critically low battery charge forces the UAS to
land in-place during the mission because it is now deemed too dangerous to make
a safe return to the charging pad. This final scenario is an important
robustness mode which reduces the risk of the UAS crashing due to a battery
malfunction.

\subsection{Long-Duration Autonomy and Recharging}

\begin{figure}%
  \centering
  \includegraphics[width=0.6\columnwidth]{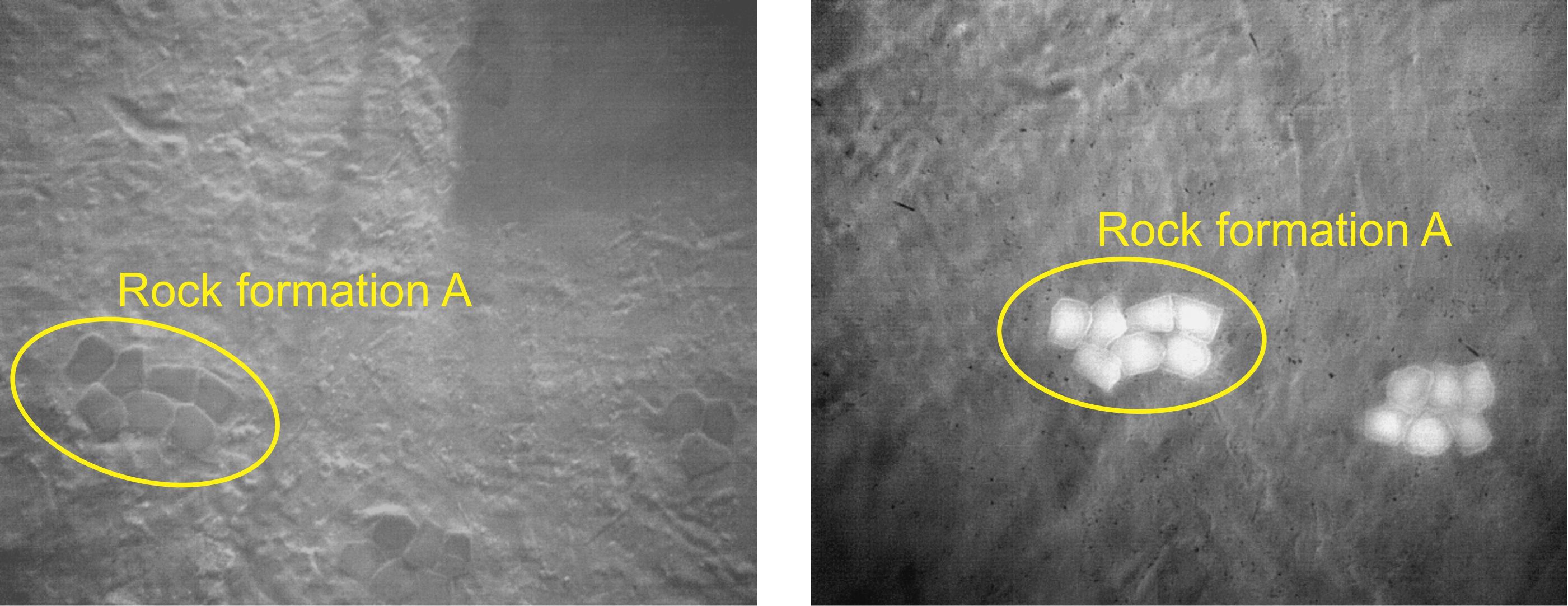}%
  \caption{Thermal images from the 4~h outdoor experiment as an example mission
    data product. Brighter pixels show higher temperatures. Left: rock formation
    in the morning. Right: same rock formation heated up at noon time.}%
  \label{fig:FLIR_image}%
\end{figure}

\begin{figure*}%
\includegraphics[width=1.0\textwidth]{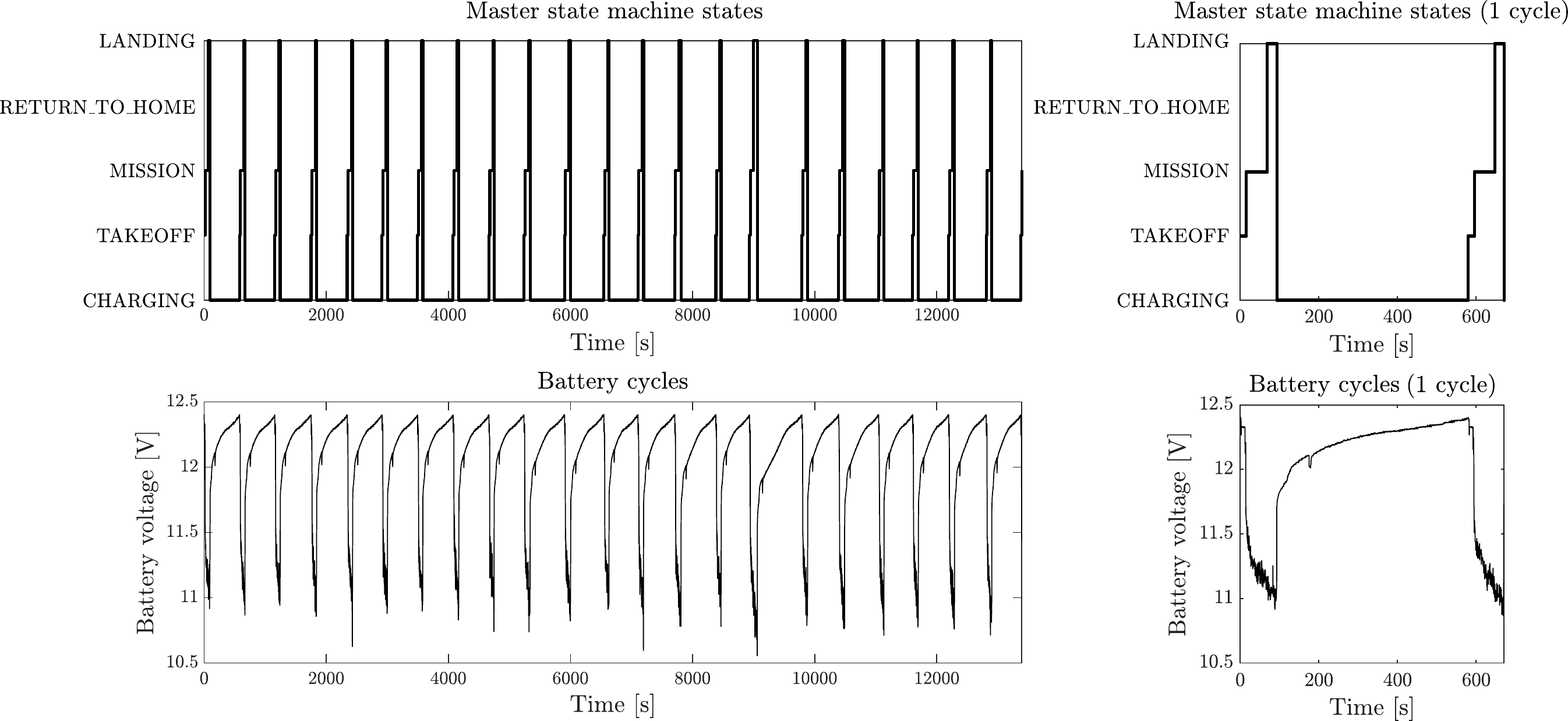}%
\caption{State transitions of the master state machine (top) and battery voltage
  level (bottom) during the 4~h outdoor experiment. Left: full
  experiment. Right: two flights.}%
\label{fig:longterm_experiment}%
\end{figure*}

\begin{figure*}%
\centering
\includegraphics[width=0.4\textwidth]{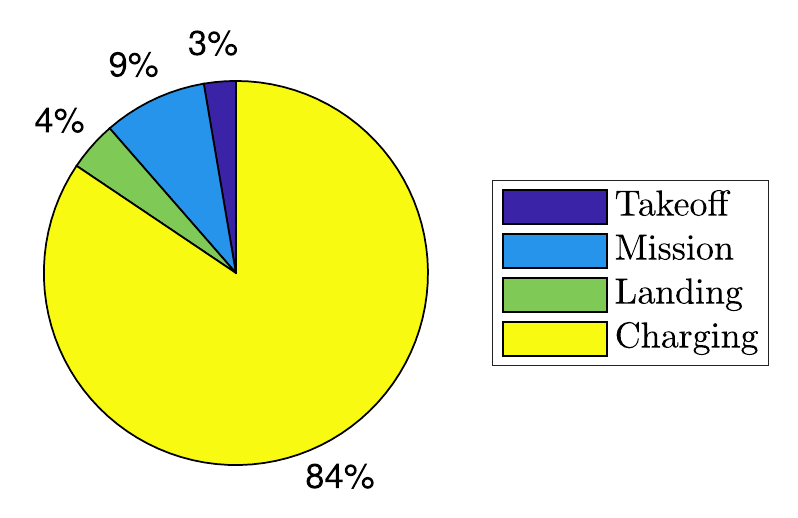}%
\caption{Breakdown of time spent in each phase of the mission during the 4~h
  outdoor experiment. About $10~\%$ of the time was spent flying in the data
  collection mode.}%
\label{fig:flighttime_breakdown}%
\end{figure*}

We tested the system during a set of indoor and outdoor deployments. For
component verification, we initially performed indoor experiments in VICON to
verify recharging only (11~h experiment with 16~flights) and then the full
system (10.6~h with 48~flights). We then proceeded to test the system outdoors
with a mission profile focused on maximizing the number of performed precision
landings. This is because precision landing is the most difficult mission phase
to execute when taking the UAS from the high-precision indoor motion capture
environment to a much lower-precision camera, IMU and GPS-based navigation
solution outdoors (see Section~\ref{sec:navigation}). The outdoor experiment was
hence a mission cycle consisting of taking off, flying to several GPS-based
waypoints for the duration of about 1~minute and then returning to the charging
pad for landing and recharging.

Among multiple outdoor deployments, our system achieved a 6~h tethered and a 4~h
free flight fully autonomous operation during which 22~flights were
performed. Figure~\ref{fig:longterm_experiment} illustrates the state
transitions of the master state machine during the free flight outdoor
experiment and the associated battery voltage reflecting the individual charging
cycles. Figure~\ref{fig:flighttime_breakdown} shows a breakdown of the times
spent in the individual mission phases, where it is seen that the UAS spent
about $10~\%$ of the time collecting data.

Although the objective of the outdoor experiment was repeatable precision
landing and not data collection, an on-board thermal camera monitored ground
surface temperatures during the course of the
experiment. Figure~\ref{fig:FLIR_image} shows an example mission data product
monitoring the surface temperature of a rock formation at different times of
day.

\subsubsection{Lessons Learned}

Outdoor operation posed several new challenges that are absent in an indoor
environment. First, dirt and dust collecting on the charging pad has at several
points prevented a proper contact with the charging pad, hence preventing
recharging. Next, camera exposure becomes a crucial factor in outdoor operation
where light conditions change throughout the day, sometimes very quickly as
clouds block sunlight. As a result, automatic exposure control is
important. Improvements can be made by adapting exposure control to the expected
brightness level of the observed landing target, or by using a different
technology like an event-based camera which has been shown to be less
susceptible to exposure effects \cite{Vidal2018}.

We also found that GPS accuracy is highly sensitive to electromagnetic
interference from nearby emitters such as Wi-Fi dongle, on-board computer and
USB connections \cite{Davuluri2013}. A careful hardware design must hence be
performed to maximize outdoor state estimation accuracy and to prevent sudden
GPS degradation due to signal interference in mid-flight.  From a software
perspective, we found the AscTec MAV framework \cite{AsctecMavFramework}, which
interfaces the AscTec autopilot board in
Figure~\ref{fig:hardware_architecture_quadrotor} to ROS, to frequently crash
after long periods of operation.

From an operational standpoint, the flight time to charging time ratio in
Figure~\ref{fig:flighttime_breakdown} is rather low. Similar ratios have been
observed before \cite{Valenti2007MissionHM}. The main limiting factors here were
the weight of the on-board power supply, which decreased flight times, and a
need to guarantee safe LiPo battery operation, which limited the charging
current and increased the charging times. A possible solution would be to
re-engineer the battery/charging pad system or to adapt a battery swapping
approach. Our software framework can easily accommodate this change.

In summary, long-term platform autonomy outdoors is hindered by minor issues
like dust accumulation, camera exposure, GPS accuracy due to signal interference
and software stability. We believe that each challenge can be solved by further
hardware and software engineering, though such fine tuning is beyond the scope
of this initial research and development. Nevertheless, due to its modular
design, we believe that the autonomy engine presented herein can readily
accommodate any such modification.

\section{Future Work}
\label{sec:discussion}

Several future improvements are conceivable for our system.  For navigation, a
rigorous analysis of latency in the data signals is necessary to improve
platform accuracy. For precision landing, AprilTag bundle pose measurement
frequency may be increased from the current 7 Hz to gather more data and
therefore to improve the estimate of the landing pad pose. To similar effect and
to increase the maximum landing approach height, the downfacing camera
resolution may be increased (e.g. \cite{Fnoop} report tag detection from a 32~m
altitude using 1080p images).

Several improvements are possible for guidance and control. Current trajectories
are simple fully constrained polynomials. To extend flight time,
energy-minimizing optimal trajectories may be generated by solving an
optimization problem. A potential avenue is to adapt existing research on using
convex optimization to plan reliable minimum-fuel trajectories for planetary
landers
\cite{Acikmese2007}.
There is also more direct research for multirotor minimum-energy and
minimum-time trajectory generation \cite{Vicencio2015,Morbidi2016,Mueller2015}.
On the control side, our experience shows that closed-loop motor RPM control can
benefit trajectory tracking accuracy as it effectively renders the motor thrust
characteristic independent of battery voltage. This makes the control system
less dependent on integrators for error correction.

Several improvements to the autonomy engine implementation are possible. More
expandable behavior can likely be achieved by porting the existing state
machines to a framework such as SMC \cite{Rapp} which follows the open-closed
design pattern of keeping software entities open for extension but closed for
modification. In practice, this means that new autonomy engine states may be
added without modifying source code for existing states. Furthermore, the
current implementation is not robust to individual ROS node crashes. This
creates single points of failure such as the AscTec HLP interface which, if it
fails, will prevent the state estimate from being sent to the HLP and will
trigger the emergency landing controller. A potential solution is to implement a
highly reliable supervisory software that is able to restart failed ROS nodes,
or the entire ROS network if necessary.

\section{Conclusion}
\label{sec:conclusion}

In this work, we presented an architecture for a long-duration fully autonomous
UAS for remote-sensing data acquisition missions. Such a system is crucial in
enabling precision agriculture and other applications that require a long-term
remote-sensing capability such as environmental monitoring and surveillance. Our
system's two major components are a fully autonomous aerial vehicle and a
landing station which integrates a charging pad. The aerial vehicle carries our
autonomy software which enables it to perpetually execute a user-defined mission
and to downlink collected data to a base station computer. We were able to
operate this system in several environments of increasing complexity,
culminating in a fully autonomous 4~h outdoor operation in which 10~\% of the
time was spent flying to collect data. This represents to the best of our
knowledge the first research publication on a UAS system that is capable of
long-duration full autonomy outdoors.

\subsubsection*{Acknowledgments}

This research was carried out at the Jet Propulsion Laboratory, California
Institute of Technology, under a contract with the National Aeronautics and
Space Administration.  The authors would like to thank Stephan Weiss for his
inputs on vehicle pose estimation in
Section~\ref{subsec:vehicle_pose_estimation} and our JPL collaborators Darren
Drewry and Debsunder Dutta for providing the NDVI data illustrated in
Section~\ref{subsec:motivation}.

Copyright 2018, California Institute of Technology. U.S. Government sponsorship
acknowledged.


\begin{thebibliography}{}

\bibitem[Achtelik et~al., 2017]{AsctecMavFramework}
Achtelik, M., Achtelik, M., Weiss, S., and Kneip, L. (2017).
\newblock asctec{\_}mav{\_}framework.
\newblock \url{http://wiki.ros.org/asctec_mav_framework}.
\newblock Accessed: 2017-12-10.

\bibitem[\Acikmese and Ploen, 2007]{Acikmese2007}
\Acikmese, B. and Ploen, S.~R. (2007).
\newblock Convex programming approach to powered descent guidance for {Mars}
  landing.
\newblock {\em {AIAA} Journal of Guidance, Control, and Dynamics},
  30(5):1353--1366.

\bibitem[Aldhaher et~al., 2017]{Aldhaher_2017}
Aldhaher, S., Mitcheson, P.~D., Arteaga, J.~M., Kkelis, G., and Yates, D.~C.
  (2017).
\newblock Light-weight wireless power transfer for mid-air charging of drones.
\newblock In {\em European Conference on Antennas and Propagation ({EUCAP})}.

\bibitem[Amazon, 2016]{Amazon2016}
Amazon (2016).
\newblock {Amazon Prime Air's} first customer delivery.
\newblock \url{https://www.youtube.com/watch?v=vNySOrI2Ny8}.
\newblock Accessed: 2017-12-11.

\bibitem[{APRIL Laboratory}, 2016]{APRILLaboratory2015}
{APRIL Laboratory} (2016).
\newblock {AprilTags} visual fiducial system.
\newblock \url{https://april.eecs.umich.edu/software/apriltag}.
\newblock Version 2015-03-18.

\bibitem[Bencina et~al., 2005]{Bencina}
Bencina, R., Kaltenbrunner, M., and Jorda, S. (2005).
\newblock Improved topological fiducial tracking in the {reacTIVision} system.
\newblock In {\em {IEEE} Computer Society Conference on Computer Vision and
  Pattern Recognition ({CVPR})}.

\bibitem[Bergamasco et~al., 2016]{Bergamasco2016}
Bergamasco, F., Albarelli, A., Cosmo, L., Rodola, E., and Torsello, A. (2016).
\newblock An accurate and robust artificial marker based on cyclic codes.
\newblock {\em {IEEE} Transactions on Pattern Analysis and Machine
  Intelligence}, 38(12):2359--2373.

\bibitem[Blackmore et~al., 2010]{Blackmore2010}
Blackmore, L., \Acikmese, B., and Scharf, D.~P. (2010).
\newblock Minimum-landing-error powered-descent guidance for {Mars} landing
  using convex optimization.
\newblock {\em {AIAA} Journal of Guidance, Control, and Dynamics},
  33(4):1161--1171.

\bibitem[Borowczyk et~al., 2017]{Borowczyk}
Borowczyk, A., Nguyen, D.-T., Nguyen, A. P.-V., Nguyen, D.~Q., Saussi{\'e}, D.,
  and Ny, J.~L. (2017).
\newblock Autonomous landing of a multirotor micro air vehicle on a high
  velocity ground vehicle.
\newblock In {\em 20th {IFAC} World Congress}.

\bibitem[Brescianini et~al., 2013]{Brescianini2013}
Brescianini, D., Hehn, M., and D'Andrea, R. (2013).
\newblock Nonlinear quadrocopter attitude control.
\newblock Technical Report 009970340, ETH Z{\"u}rich, Departement Maschinenbau
  und Verfahrenstechnik.

\bibitem[Brockers et~al., 2011]{Brockers2011}
Brockers, R., Bouffard, P., Ma, J., Matthies, L., and Tomlin, C. (2011).
\newblock Autonomous landing and ingress of micro-air-vehicles in urban
  environments based on monocular vision.
\newblock In {\em Micro- and Nanotechnology Sensors, Systems, and Applications
  {III}}.

\bibitem[Brockers et~al., 2012]{Brockers2012}
Brockers, R., Susca, S., Zhu, D., and Matthies, L. (2012).
\newblock Fully self-contained vision-aided navigation and landing of a micro
  air vehicle independent from external sensor inputs.
\newblock In {\em {SPIE} Unmanned Systems Technology {XIV}}.

\bibitem[Brommer et~al., 2018]{Brommer2018}
Brommer, C., Malyuta, D., Hentzen, D., and Brockers, R. (2018).
\newblock Long-duration autonomy for small rotorcraft {UAS} including
  recharging.
\newblock In {\em {IEEE/RSJ} International Conference on Intelligent Robots and
  Systems ({IROS})}.

\bibitem[Burri et~al., 2015]{Burri2015}
Burri, M., Oleynikova, H., Achtelik, M.~W., and Siegwart, R. (2015).
\newblock Real-time visual-inertial mapping, re-localization and planning
  onboard {MAVs} in unknown environments.
\newblock In {\em {IEEE/RSJ} International Conference on Intelligent Robots and
  Systems ({IROS})}.

\bibitem[Chaves et~al., 2015]{Chaves2015}
Chaves, S.~M., Wolcott, R.~W., and Eustice, R.~M. (2015).
\newblock {NEEC} research: Toward {GPS}-denied landing of unmanned aerial
  vehicles on ships at sea.
\newblock {\em Naval Engineers Journal}, 127(1):23--35.

\bibitem[Davuluri and Chen, 2013]{Davuluri2013}
Davuluri, P. and Chen, C. (2013).
\newblock Radio frequency interference due to {USB}3 connector radiation.
\newblock In {\em {IEEE} International Symposium on Electromagnetic
  Compatibility}.

\bibitem[de~Almeida and Akella, 2017]{DeAlmeida2017}
de~Almeida, M.~M. and Akella, M. (2017).
\newblock New numerically stable solutions for minimum-snap quadcopter
  aggressive maneuvers.
\newblock In {\em American Control Conference ({ACC})}.

\bibitem[Devernay and Faugeras, 2001]{Devernay}
Devernay, F. and Faugeras, O. (2001).
\newblock Straight lines have to be straight.
\newblock {\em Machine Vision and Applications}, 13(1):14--24.

\bibitem[Faessler et~al., 2017]{Faessler2016}
Faessler, M., Falanga, D., and Scaramuzza, D. (2017).
\newblock Thrust mixing, saturation, and body-rate control for accurate
  aggressive quadrotor flight.
\newblock {\em {IEEE} Robotics and Automation Letters}, 2(2):476--482.

\bibitem[Faessler et~al., 2015]{Faessler2015}
Faessler, M., Fontana, F., Forster, C., and Scaramuzza, D. (2015).
\newblock Automatic re-initialization and failure recovery for aggressive
  flight with a monocular vision-based quadrotor.
\newblock In {\em {IEEE} International Conference on Robotics and Automation
  ({ICRA})}.

\bibitem[Fiala, 2005]{Fiala}
Fiala, M. (2005).
\newblock {ARTag}, a fiducial marker system using digital techniques.
\newblock In {\em {IEEE} Computer Society Conference on Computer Vision and
  Pattern Recognition ({CVPR})}.

\bibitem[{Food and Agriculture Organization of the United Nations},
  2009]{FAOFood}
{Food and Agriculture Organization of the United Nations} (2009).
\newblock 2050: A third more mouths to feed.
\newblock \url{http://www.fao.org/news/story/en/item/35571/icode/}.
\newblock Accessed: 2017-12-10.

\bibitem[Forster et~al., 2015]{Forstera}
Forster, C., Faessler, M., Fontana, F., Werlberger, M., and Scaramuzza, D.
  (2015).
\newblock Continuous on-board monocular-vision-based elevation mapping applied
  to autonomous landing of micro aerial vehicles.
\newblock In {\em {IEEE} International Conference on Robotics and Automation
  ({ICRA})}.

\bibitem[Forster et~al., 2014]{Forster}
Forster, C., Pizzoli, M., and Scaramuzza, D. (2014).
\newblock {SVO}: Fast semi-direct monocular visual odometry.
\newblock In {\em {IEEE} International Conference on Robotics and Automation
  ({ICRA})}.

\bibitem[Furrer et~al., 2016]{rotors:2016}
Furrer, F., Burri, M., Achtelik, M., and Siegwart, R. (2016).
\newblock {RotorS} -- a modular gazebo {MAV} simulator framework.
\newblock In Koubaa, A., editor, {\em Robot Operating System ({ROS}) The
  Complete Reference}, volume~1, chapter~23, pages 595--625. Springer.

\bibitem[{Goldman Sachs}, 2017]{GoldmanSachs}
{Goldman Sachs} (2017).
\newblock Drones: Reporting for work.
\newblock
  \url{http://www.goldmansachs.com/our-thinking/technology-driving-innovation/drones/}.
\newblock Accessed: 2017-12-10.

\bibitem[GoodRobots, 2019]{Fnoop}
GoodRobots (2019).
\newblock vision{\_}landing.
\newblock \url{https://github.com/goodrobots/vision_landing}.
\newblock Accessed: 2019-05-01.

\bibitem[Grewal et~al., 2007]{Grewal2007}
Grewal, M.~S., Weill, L.~R., and Andrews, A.~P. (2007).
\newblock {\em Global Positioning Systems, Inertial Navigation, and
  Integration}.
\newblock Wiley, Hoboken, NJ.

\bibitem[Hast et~al., 2013]{Hast2013}
Hast, A., Nysj{\"o}, J., and Marchetti, A. (2013).
\newblock Optimal {RANSAC} -- towards a repeatable algorithm for finding the
  optimal set.
\newblock {\em Journal of WSCG}, 21(1):21--30.

\bibitem[Jin et~al., 2017]{Jin2017}
Jin, P., Matikainen, P., and Srinivasa, S.~S. (2017).
\newblock Sensor fusion for fiducial tags: highly robust pose estimation from
  single frame {RGBD}.
\newblock In {\em {IEEE}/{RSJ} International Conference on Intelligent Robots
  and Systems ({IROS})}.

\bibitem[Jin et~al., 2016]{Shaogang2016}
Jin, S., Zhang, J., Shen, L., and Li, T. (2016).
\newblock On-board vision autonomous landing techniques for quadrotor: a
  survey.
\newblock In {\em Chinese Control Conference ({CCC})}.

\bibitem[Klein and Murray, 2007]{Klein2007}
Klein, G. and Murray, D. (2007).
\newblock Parallel tracking and mapping for small {AR} workspaces.
\newblock In {\em {IEEE} and {ACM} International Symposium on Mixed and
  Augmented Reality}.

\bibitem[Koenig and Howard, 2004]{Koenig}
Koenig, N. and Howard, A. (2004).
\newblock {Design and use paradigms for {Gazebo}, an open-source multi-robot
  simulator}.
\newblock In {\em IEEE/RSJ International Conference on Intelligent Robots and
  Systems (IROS)}.

\bibitem[Kyristsis et~al., 2016]{Kyristsis2016}
Kyristsis, S., Antonopoulos, A., Chanialakis, T., Stefanakis, E., Linardos, C.,
  Tripolitsiotis, A., and Partsinevelos, P. (2016).
\newblock Towards autonomous modular {UAV} missions: the detection,
  geo-location and landing paradigm.
\newblock {\em Sensors}, 16(11):1844.

\bibitem[Ling, 2014]{Ling2014}
Ling, K. (2014).
\newblock Precision landing of a quadrotor {UAV} on a moving target using
  low-cost sensors.
\newblock Master's thesis, University of Waterloo, Waterloo.

\bibitem[Lupashin et~al., 2014]{Lupashin2014}
Lupashin, S., Hehn, M., Mueller, M.~W., Schoellig, A.~P., Sherback, M., and
  D'Andrea, R. (2014).
\newblock A platform for aerial robotics research and demonstration: the flying
  machine arena.
\newblock {\em Mechatronics}, 24(1):41--54.

\bibitem[Malyuta, 2018]{Malyuta2018}
Malyuta, D. (2018).
\newblock Guidance, navigation, control and mission logic for quadrotor
  full-cycle autonomy.
\newblock Master's thesis, ETH Z{\"u}rich, Z{\"u}rich.

\bibitem[Mellinger and Kumar, 2011]{Mellinger2011}
Mellinger, D. and Kumar, V. (2011).
\newblock Minimum snap trajectory generation and control for quadrotors.
\newblock In {\em {IEEE} International Conference on Robotics and Automation
  (ICRA)}.

\bibitem[Morbidi et~al., 2016]{Morbidi2016}
Morbidi, F., Cano, R., and Lara, D. (2016).
\newblock Minimum-energy path generation for a quadrotor {UAV}.
\newblock In {\em {IEEE} International Conference on Robotics and Automation
  ({ICRA})}.

\bibitem[Mueller and D'Andrea, 2012]{Mueller2012}
Mueller, M.~W. and D'Andrea, R. (2012).
\newblock Critical subsystem failure mitigation in an indoor {UAV} testbed.
\newblock In {\em {IEEE}/{RSJ} International Conference on Intelligent Robots
  and Systems (IROS)}.

\bibitem[Mueller and D'Andrea, 2013]{Mueller2013}
Mueller, M.~W. and D'Andrea, R. (2013).
\newblock A model predictive controller for quadrocopter state interception.
\newblock In {\em European Control Conference ({ECC})}.

\bibitem[Mueller et~al., 2015]{Mueller2015}
Mueller, M.~W., Hehn, M., and D'Andrea, R. (2015).
\newblock A computationally efficient motion primitive for quadrocopter
  trajectory generation.
\newblock {\em {IEEE} Transactions on Robotics}, 31(6):1294--1310.

\bibitem[Mulgaonkar and Kumar, 2014]{Mulgaonkar_2014}
Mulgaonkar, Y. and Kumar, V. (2014).
\newblock Autonomous charging to enable long-endurance missions for small
  aerial robots.
\newblock In {\em Micro- and Nanotechnology Sensors, Systems, and Applications
  {VI}}.

\bibitem[Nissler et~al., 2016]{Nissler2016}
Nissler, C., Buttner, S., Marton, Z.-C., Beckmann, L., and Thomasy, U. (2016).
\newblock Evaluation and improvement of global pose estimation with multiple
  {AprilTags} for industrial manipulators.
\newblock In {\em {IEEE} International Conference on Emerging Technologies and
  Factory Automation ({ETFA})}.

\bibitem[Olson, 2011]{olson2011tags}
Olson, E. (2011).
\newblock {AprilTag}: A robust and flexible visual fiducial system.
\newblock In {\em {IEEE} International Conference on Robotics and Automation
  (ICRA)}.

\bibitem[{OpenCV}, 2017]{OpenCV}
{OpenCV} (2017).
\newblock {cv::solvePnP()}.
\newblock
  \url{https://docs.opencv.org/3.3.1/d9/d0c/group__calib3d.html#ga549c2075fac14829ff4a58bc931c033d}.
\newblock Accessed: 2017-12-10.

\bibitem[Rapp et~al., 2017]{Rapp}
Rapp, C., Suez, E., Perrad, F., Liscio, C., Arnold, T., and Krajcovic, G.
  (2017).
\newblock {SMC}: The state machine compiler.
\newblock \url{http://smc.sourceforge.net/}.
\newblock Accessed: 2017-12-10.

\bibitem[Richter et~al., 2016]{Richter2013}
Richter, C., Bry, A., and Roy, N. (2016).
\newblock Polynomial trajectory planning for aggressive quadrotor flight in
  dense indoor environments.
\newblock In {\em Springer Tracts in Advanced Robotics}, volume 114, pages
  649--666. Springer.

\bibitem[Rublee et~al., 2011]{Rublee}
Rublee, E., Rabaud, V., Konolige, K., and Bradski, G. (2011).
\newblock {ORB}: An efficient alternative to {SIFT} or {SURF}.
\newblock In {\em {IEEE} International Conference on Computer Vision (ICCV)}.

\bibitem[Skogestad and Postlethwaite, 2005]{Skogestad2005}
Skogestad, S. and Postlethwaite, I. (2005).
\newblock {\em Multivariable feedback control: analysis and design}.
\newblock Wiley, 2 edition.

\bibitem[{Skysense}, 2018]{skysense_2017}
{Skysense} (2018).
\newblock {Skysense} charging pad.
\newblock \url{https://www.skysense.co/charging-pad-outdoor}.
\newblock Accessed: 2018-10-08.

\bibitem[Suzuki et~al., 2011]{Suzuki_2012}
Suzuki, K. A.~O., Filho, P.~K., and Morrison, J.~R. (2011).
\newblock Automatic battery replacement system for {UAVs}: analysis and design.
\newblock {\em Journal of Intelligent {\&} Robotic Systems}, 65(1-4):563--586.

\bibitem[Toksoz et~al., 2011]{Toksoz_2011}
Toksoz, T., Redding, J., Michini, M., Michini, B., How, J., Vavrina, M., and
  Vian, J. (2011).
\newblock Automated battery swap and recharge to enable persistent {UAV}
  missions.
\newblock In {\em AIAA Infotech@Aerospace}.

\bibitem[Valenti et~al., 2007]{Valenti2007MissionHM}
Valenti, M., Dale, D., How, J., de~Farias, D.~P., and Vian, J. (2007).
\newblock Mission health management for 24/7 persistent surveillance
  operations.
\newblock In {\em {AIAA} Guidance, Navigation and Control Conference and
  Exhibit}.

\bibitem[Vicencio et~al., 2015]{Vicencio2015}
Vicencio, K., Korras, T., Bordignon, K.~A., and Gentilini, I. (2015).
\newblock Energy-optimal path planning for six-rotors on multi-target missions.
\newblock In {\em {IEEE}/{RSJ} International Conference on Intelligent Robots
  and Systems ({IROS})}.

\bibitem[Vidal et~al., 2018]{Vidal2018}
Vidal, A.~R., Rebecq, H., Horstschaefer, T., and Scaramuzza, D. (2018).
\newblock Ultimate {SLAM}? combining events, images, and {IMU} for robust
  visual {SLAM} in {HDR} and high-speed scenarios.
\newblock {\em {IEEE} Robotics and Automation Letters}, 3(2):994--1001.

\bibitem[Vincent, 2017]{Vincent2017}
Vincent, J. (2017).
\newblock {Google's Project Wing has successfully tested its air traffic
  control system for drones}.
\newblock
  \url{https://www.theverge.com/2017/6/8/15761220/google-project-wing-drone-air-traffic-control-tests}.
\newblock Accessed: 2017-12-10.

\bibitem[Wang and Olson, 2016]{wang2016iros}
Wang, J. and Olson, E. (2016).
\newblock {AprilTag} 2: Efficient and robust fiducial detection.
\newblock In {\em {IEEE}/{RSJ} International Conference on Intelligent Robots
  and Systems ({IROS})}.

\bibitem[Weiss et~al., 2012]{Weiss2012}
Weiss, S., Achtelik, M.~W., Chli, M., and Siegwart, R. (2012).
\newblock Versatile distributed pose estimation and sensor self-calibration for
  an autonomous {MAV}.
\newblock In {\em {IEEE} International Conference on Robotics and Automation
  (ICRA)}.

\bibitem[Wingtra, 2017]{Wingtra}
Wingtra (2017).
\newblock How it works.
\newblock \url{https://wingtra.com/workflow/}.
\newblock Accessed: 2017-12-10.

\bibitem[Xu and Dudek, 2011]{Xu2011}
Xu, A. and Dudek, G. (2011).
\newblock Fourier tag: A smoothly degradable fiducial marker system with
  configurable payload capacity.
\newblock In {\em Canadian Conference on Computer and Robot Vision}.

\bibitem[Yang et~al., 2012]{Yang2013}
Yang, S., Scherer, S.~A., and Zell, A. (2012).
\newblock An onboard monocular vision system for autonomous takeoff, hovering
  and landing of a micro aerial vehicle.
\newblock {\em Journal of Intelligent {\&} Robotic Systems}, 69(1-4):499--515.

\end{thebibliography}

\end{document}